\documentclass[11pt]{article}

\usepackage{multirow}
\usepackage{booktabs}
\usepackage[table]{xcolor}
\definecolor{best}{HTML}{9DC2E8}
\definecolor{second}{HTML}{E6F0FA}
\usepackage{booktabs}
\usepackage{multirow}
\usepackage{array}
\usepackage{xcolor}
\usepackage{colortbl}
\usepackage{siunitx}
\usepackage{makecell}
\definecolor{headerblue}{RGB}{220,230,242}
\definecolor{overallgray}{RGB}{240,240,240}
\definecolor{colorgreen}{HTML}{43A047}
\usepackage{graphicx}
\usepackage{listings}
\usepackage{xcolor}
\usepackage{algorithm}
\usepackage{algorithmic}
\usepackage{amsmath}
\usepackage{amssymb}
\usepackage{tcolorbox}
\usepackage{enumitem}
\usepackage{float}
\newtcolorbox{promptbox}[1][]{
  colback=gray!5,
  colframe=gray!50,
  fonttitle=\bfseries,
  title={#1},
  boxrule=0.5pt,
  left=6pt, right=6pt, top=4pt, bottom=4pt,
  fontupper=\small,
  breakable,
}

\tcbuselibrary{listings, breakable}

\lstdefinestyle{promptcode}{
  language=Python,
  basicstyle=\ttfamily,
  keywordstyle=\bfseries,
  breaklines=true,
  columns=fullflexible,
  aboveskip=2pt, belowskip=2pt
}

\newcommand{\logline}[1]{{\ttfamily #1}}

\usepackage[final]{acl}

\usepackage{times}
\usepackage{latexsym}

\usepackage[T1]{fontenc}

\usepackage[utf8]{inputenc}

\usepackage{microtype}

\usepackage{inconsolata}

\usepackage{graphicx}

\title{MedFeat: Model-Aware and Explainability-Driven Feature Engineering
with LLMs for Tabular Prediction}

\author{
 \textbf{Zizheng Zhang\textsuperscript{1,2}},
 \textbf{Yiming Li\textsuperscript{1}},
 \textbf{Justin Xu\textsuperscript{1}},
 \textbf{Jinyu Wang\textsuperscript{2}},
 \textbf{Rui Wang\textsuperscript{2}},
\\
 \textbf{Lei Song\textsuperscript{2}},
 \textbf{Jiang Bian\textsuperscript{2}},
 \textbf{David W Eyre\textsuperscript{1}},
 \textbf{Jingjing Fu\textsuperscript{2}}
\\
\\
 \textsuperscript{1}University of Oxford,
 \textsuperscript{2}Microsoft Research
\\
}

\begin{document}
\maketitle

\begin{abstract}

In clinical tabular prediction, classical machine learning models with feature engineering often outperform neural methods. LLMs are increasingly used to automate this process, acting as domain experts that propose diverse feature transformations to boost downstream performance. However, the feature generation process of existing LLM-based methods is agnostic to the downstream learner: the LLM receives no signal about which features currently drive predictions or where the model's representational capacity falls short, so proposals are neither targeted to promising regions of the feature space nor tailored to the learner's inductive bias. This shortcoming is amplified in healthcare data, which simultaneously exhibits class imbalance, heterogeneous feature spaces, and strict interpretability requirements. In this paper, we propose MedFeat, the first feature engineering framework inspired by the workflow of machine learning practitioners, leveraging model-awareness and feature importance signals to iteratively guide feature discovery for clinical tabular learning. We evaluate MedFeat on a broad range of challenging real-world clinical tasks and show that it statistically significantly outperforms state-of-the-art baselines, with an average F1 improvement of more than 10\% over the baseline across models with distinct inductive biases.

\end{abstract}

\section{Introduction}

Clinical prediction models support numerous high-stakes decisions in healthcare, including early warning for deterioration, surveillance, and discharge prediction\cite{ bonnett2019guide, smith2019national, zhou2023improving}. Real-world healthcare datasets are difficult to model reliably due to severe class imbalance, heterogeneous feature types, and complex temporal patterns. While deep learning excels on imaging, tree-based models often outperform it on tabular tasks \cite{zhang2026less, grinsztajn2022tree}, and deployment is further constrained by interpretability, privacy, and training cost. As a result, classical models combined with feature engineering remain widely favoured in practice \cite{liao2022does}.

\begin{figure*}[h]
  \centering
  \includegraphics[width=1\textwidth]{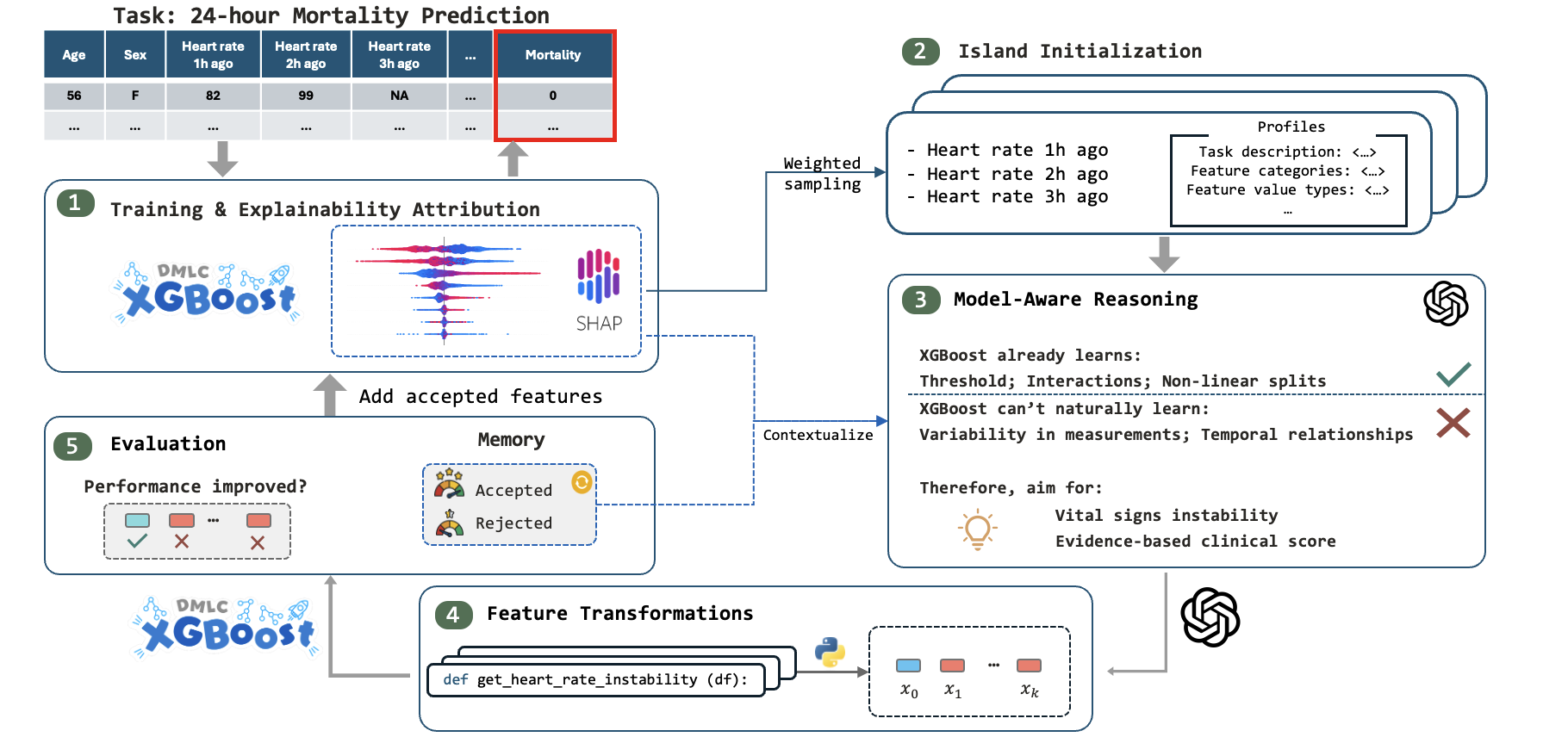}
  \caption{\textbf{Overview of MedFeat.} MedFeat iteratively augments clinical tabular data with LLM-generated features. (1) Starting from a model trained on raw features, the framework computes feature importance. (2) It then forms importance-weighted “feature islands” by sampling a subset of influential features, (3) and guides the LLM with the island profiles, model constraints, and a memory bank to propose clinically meaningful transformations. (4) Proposed features are validated and evaluated locally, (5) accepted features are added to the data, and the feedback memory is updated for the next iteration. }
  \label{fig:medfeat}
\end{figure*}

Manual feature engineering is labor-intensive and requires substantial domain expertise. Automated methods such as AutoFeat and OpenFE \cite{horn2019autofeat, zhang2023openfe} address this by exhaustively combining transformations from pre-defined operators, but produce unwieldy feature spaces and lack task-specific context.

Large Language Models (LLMs) offer new opportunities for feature engineering: they encode rich prior domain knowledge and possess compositional reasoning capabilities. Recent work in this direction includes CAAFE \cite{hollmann2023large}, FeatLLM \cite{han2024large}, and OCTree \cite{nam2024optimized}, which synthesize threshold rules, binary discriminative features, or decision-tree-based feedback loops to guide LLM-driven feature generation.

Although these methods demonstrate the feasibility of LLMs contributing useful features, current feature-generation workflows have been constructed in a largely intuitive manner, detached from the informed reasoning process that domain practitioners naturally follow. None of the existing methods explicitly accounts for the two basic concepts in tabular learning: the model type and the feature attribution. Almost all existing approaches treat features as interchangeable and remain downstream-learner-agnostic \cite{nam2024optimized}. By contrast, a practitioner inherently adapts to the learner's representational capacity and prioritizes features by their predictive relevance before proposing targeted transformations \cite{kuhn2019feature, duboue2020art}. As a result, existing methods may target low-relevance features or produce transformations that the model can already partially or fully recover from its own structure during training, while overlooking transformations that would genuinely expand the model’s representational capacity.

\paragraph{Contributions} In this paper, we propose \textbf{MedFeat}, a \textbf{M}odel-aware and \textbf{E}xplainability-\textbf{D}riven agentic \textbf{Feat}ure engineering framework inspired by machine learning practitioners. We abstract four empirical operational signals that constitute a feature engineering workflow: 
(1) \textit{Domain knowledge}: what transformations are clinically meaningful? 
(2) \textit{Model-awareness}: What patterns has the downstream model already learned during training? 
(3) \textit{Feature Attribution}: Where in the current feature space does the predictive signal lie? 
(4) \textit{Operation record}: Which transformations have proven useful or unsuccessful? \cite{kuhn2019feature}
Together, these four signals reflect the established iterative workflow of model development in practice: assess the model, understand its attribution, find the signal, propose a change, record the outcome, and repeat. We formalise this integrated view as the basis for MedFeat. To the best of our knowledge, MedFeat is the first feature engineering framework that mirrors the cognitive process of machine learning practitioners, and the first to leverage the downstream learner’s inductive bias together with feature importance to guide feature discovery.

We evaluate MedFeat on a broad suite of clinical tabular prediction tasks spanning multiple datasets and outcomes, including settings with severe class imbalance and heterogeneous missingness, where most existing approaches fail. Our experiments show that MedFeat consistently improves the performance of strong baselines, proposes clinically meaningful features, and remains compatible with practical clinical constraints. We also conduct a simulated deployment study showing that MedFeat-generated features remain useful in different clinical environments.

\section{Related Work}

\paragraph{Machine Learning for Healthcare.} Advances in machine learning and the increasing availability of hospital data have led to a wide range of approaches for processing patient data for clinical prediction \cite{ kirchler2026large}. Deep learning has achieved remarkable success in computer
vision \cite{krizhevsky2012imagenet} and has been extended
to other healthcare modalities such as imaging and free text,
with applications including tumor segmentation and information
extraction from clinical notes \cite{ronneberger2015u, isensee2021nnu, alsentzer2019publicly}. However, it often underperforms on structured tabular prediction tasks, especially when the data contains mixed feature types \cite{grinsztajn2022tree}. Compared with curated machine learning benchmarks, healthcare tabular data are typically noisy and simultaneously exhibit multiple complex patterns. These properties make generalization difficult, particularly under temporal drift and cross-site distribution shifts \cite{guan2025keeping}. In this context, classical models such as XGBoost remain strong and are commonly deployed because of their robustness, ease of optimization, and interpretability. In this paper, we explore whether the performance of such classical models can be further boosted by our framework.

\paragraph{Autonomous Feature Engineering.} Automatic feature engineering reduces manual feature construction by generating candidate transformations from a fixed operator library and performing post-hoc selection. Methods such as AutoFeat enumerate feature compositions (e.g., nonlinear transforms and pairwise interactions) and apply selection to control redundancy \cite{horn2019autofeat}. OpenFE follows a similar operator-driven design but improves scalability via pruning and staged evaluation \cite{zhang2023openfe}. While effective in some settings, these approaches are limited for clinical tabular data: the search space is constrained by generic operators, the feature space grows exponentially under exhaustive combination, and validation-score-based selection is unstable under label noise and severe class imbalance. 

LLM-based feature engineering replaces fixed operator libraries with LLM-proposed rules or code, leveraging the LLM’s language understanding and reasoning capabilities. CAAFE was the first LLM-based framework to incorporate domain knowledge into automated feature engineering by passing dataset descriptions to an LLM, and it iterates by retaining features that improve validation performance \cite{hollmann2023large}. FeatLLM leverages language models to synthesize class-discriminative features from few-shot samples and ensembles them into binary features through LLM reasoning \cite{han2024large}. Similarly, OCTree queries an LLM to synthesize a single discriminative rule at each iteration and refines rules using decision-tree-based reasoning feedback from prior trials to maximize the validation score \cite{nam2024optimized}. Unlike previous approaches, our framework is motivated by the practice of domain experts and aims to guide feature discovery more effectively and robustly.

\section{Model-aware and Explainability-Driven Feature Engineering}
We formally formulate the research question and introduce our framework, MedFeat. MedFeat uses model explainability signals to obtain informative optimization signals and explicitly conditions feature generation on the inductive bias of the downstream learner.

\paragraph{Research Question Formulation} We focus on supervised clinical tabular prediction. Let \(D = \{(x_i, y_i)\}_{i=1}^{N}\) denote a dataset of \(N\) patient episodes, where each input
\(x_i \in \mathcal{X} \subseteq \mathbb{R}^{M}\)
is an \(M\)-dimensional column feature vector and
\(y_i \in \mathcal{Y}\)
is the corresponding label. We denote the set of column names by
\(C = \{c_1, c_2, \ldots, c_M\}\),
so that \(x_{i,j}\) corresponds to column \(c_j\). In this paper, we focus on binary classification tasks with \(\mathcal{Y} = \{0,1\}\); though the framework extends naturally to regression.
Given a training set $\mathcal{D}_{\text{train}}$ and a validation set $\mathcal{D}_{\text{val}}$, 
the objective is to find a set of feature transformations 
$\boldsymbol{\sigma} = \{\sigma_1, \ldots, \sigma_K\}$ 
that maps the original feature space to an augmented representation and maximizes validation performance.
Each transformation $\sigma_k : \mathbb{R}^{M} \rightarrow \mathbb{R}$ produces a new one-dimensional
feature from the current representation. Applying $\boldsymbol{\sigma}$ to the dataset yields the
augmented dataset
\[
\mathcal{D} \oplus \boldsymbol{\sigma}
=
\{(x_i \oplus \sigma_1(x_i) \oplus \cdots \oplus \sigma_K(x_i), y_i)\}_{i=1}^{N}.
\]
Let $f_\theta$ denote a downstream model and let
$\mathcal{L}(\cdot)$ denote the evaluation metric of the given prediction task.
We formalize feature engineering as the following optimization problem:

\begin{equation}
\begin{aligned}
& \max_{\boldsymbol{\sigma} \in \Sigma}
\; \mathcal{L}\!\left(f_{\boldsymbol{\sigma}}^{\star},\, \mathcal{D}_{\text{val}} \oplus \boldsymbol{\sigma}\right) \\
& \text{s.t. } f_{\boldsymbol{\sigma}}^{\star}
= \arg\max_{f}
\mathcal{L}\!\left(f,\, \mathcal{D}_{\text{train}} \oplus \boldsymbol{\sigma}\right)
\end{aligned}
\end{equation}

where $\Sigma$ denotes the space of admissible feature transformations.
This objective seeks transformations that yield maximal performance on validation data \cite{nam2024optimized}.

Directly solving this optimization is computationally intractable due to the non-differentiable and
combinatorial nature of $\Sigma$, as well as the need to repeatedly retrain $f$ for each candidate
transformation set. Consequently, MedFeat approaches this problem via iterative, explainability-driven
optimization, using structured feedback from the downstream model to guide the search over
feature transformations.

\subsection{MedFeat Overview}

MedFeat proceeds iteratively for $T$ rounds. After training an initial baseline model on $\mathcal{D}_{\text{train}}$, each iteration $t$ maintains a set of accepted transformations $\boldsymbol{\sigma}_t = \{\sigma_1, \ldots, \sigma_{t-1}\}$ and a corresponding downstream model $f_{\theta_t}$ trained on the augmented dataset $\mathcal{D}_{\text{train}} \oplus \boldsymbol{\sigma}_t$. We compute feature importance from $f_{\theta_t}$ using SHapley Additive exPlanations (SHAP) values applied to $\mathcal{D}_{\text{val}}$ \cite{lundberg2017unified}, which defines a probability distribution over features used to construct $K$ candidate \emph{islands}-subsets of features that serve as feature-generation contexts. For each island, MedFeat builds a structured model-aware prompt and queries an LLM for a candidate feature transformation, which is then applied and evaluated on the augmented dataset. The best-performing island is accepted if it exceeds the current baseline by a tolerance threshold (introduced to account for noise in the validation set); otherwise all candidates from the iteration are rejected. The baseline model and feature importance scores are then updated, and the loop repeats until $T$ rounds complete (Figure~\ref{fig:medfeat}, Appendix~\ref{app:algorithm}).

\subsection{Explainability-Driven Search}
Let $f_{\theta_t}$ denote the baseline model at iteration $t$. We compute a feature relevance vector $s_t$ from $f_{\theta_t}$ using SHAP values on $\mathcal{D}_{\text{val}}$, which serves as the primary feedback signal in MedFeat. These scores rank features by importance, guide island sampling, and track how feature relevance evolves across iterations.

At each iteration, MedFeat leverages $s_t$ to construct $K$ islands, where each island is a subset of $m$ features sampled from the current feature space according to type and importance. Let $\tilde{s}_{t,i}$ denote the normalized importance of the $i$-th feature group; features are sampled from $p_t(i) = \tilde{s}_{t,i}$, balancing exploitation of highly relevant features with exploration of less prominent ones. Restricting each LLM call to a small island focuses the LLM on localized regions of the feature space, while multiple islands per iteration enable parallel exploration of diverse feature combinations. Further implementation details are provided in Appendix~\ref{app:implementations}.

\begin{table*}[h]
\centering
\resizebox{\textwidth}{!}{%
\setlength{\tabcolsep}{4pt}
\newcommand{\std}[1]{{\scriptsize$\pm$#1}}
\begin{tabular}{@{}c*{7}{cc}c@{}}
\toprule
\multirow{2}{*}{\raisebox{-0.3em}{\textbf{Method}}}
 & \multicolumn{2}{c}{\textbf{Mortality-ICU}}
 & \multicolumn{2}{c}{\textbf{Mortality-10yrs}}
 & \multicolumn{2}{c}{\textbf{Heart Failure}}
 & \multicolumn{2}{c}{\textbf{Mortality-Ward}}
 & \multicolumn{2}{c}{\textbf{Cog-Impairment}}
 & \multicolumn{2}{c}{\textbf{Discharge}}
 & \multicolumn{2}{c}{\textbf{Readmission}}
 & \multirow{2}{*}{\makecell{\textbf{Avg.} \\ \textbf{Rank $\downarrow$}}} \\
\cmidrule(lr){2-3} \cmidrule(lr){4-5} \cmidrule(lr){6-7} \cmidrule(lr){8-9} \cmidrule(lr){10-11} \cmidrule(lr){12-13} \cmidrule(lr){14-15}
 & F1 & AUROC & F1 & AUROC & F1 & AUROC & F1 & AUROC & F1 & AUROC & F1 & AUROC & F1 & AUROC &  \\

\midrule
\multicolumn{16}{@{}l}{\textit{XGBoost}} \\

Baseline 
& 9.5\std{0.9} & 75.6\std{1.3} 
& 23.6\std{0.8} & 68.3\std{1.9} 
& 12.9\std{0.6} & 67.4\std{1.2} 
& 1.4\std{0.2} & 71.0\std{1.4} 
& 55.3\std{1.7} & 73.8\std{1.6} 
& 39.8\std{0.4} & 71.0\std{0.5} 
& 29.7\std{0.5} & 58.7\std{1.7} 
& 4.1 \\

OpenFE 
& 10.2\std{0.5} & 73.8\std{0.5} 
& 25.0\std{1.8} & \cellcolor{second} 69.2\std{1.9} 
& 12.2\std{0.4} & 65.6\std{1.0} 
& \cellcolor{best} \textbf{6.1}\std{4.1} & \cellcolor{second} 74.8\std{1.1} 
& \cellcolor{second} 56.7\std{1.1} & \cellcolor{second} 74.7\std{1.0} 
& 38.1\std{0.8} & 68.9\std{1.1} 
& 29.2\std{0.3} & 59.0\std{0.4} 
& 3.6 \\

CAAFE 
& 10.0\std{0.7} & 75.8\std{1.4} 
& 24.9\std{1.6} & 68.2\std{2.1} 
& 12.8\std{0.6} & \cellcolor{second} 67.9\std{0.7} 
& 1.5\std{0.5} & 68.3\std{3.2} 
& 55.6\std{1.3} & 74.1\std{1.4} 
& 40.2\std{0.8} & \cellcolor{second} 71.1\std{0.7} 
& 30.0\std{0.2} & \cellcolor{second} 60.4\std{0.7} 
& 3.2 \\

FeatLLM 
& \small{N/A} & 50.0\std{0.0} 
& \small{N/A} & 55.9\std{5.5} 
& \small{N/A} & 50.0\std{0.0} 
& \small{N/A} & 50.0\std{0.0} 
& \small{N/A} & 51.3\std{1.9} 
& \small{N/A} & 50.0\std{0.0} 
& \small{N/A} & 50.0\std{0.0} 
& 6.0 \\

OCTree 
& \cellcolor{second} 10.6\std{1.7} & \cellcolor{second} 76.2\std{1.2}
& \cellcolor{second} 25.1\std{1.5} & 68.2\std{2.3} 
& \cellcolor{second} 13.0\std{0.4} & \cellcolor{second} 67.9\std{0.7} 
& 1.7\std{0.1} & 69.9\std{2.5} 
& 56.5\std{1.4} & 74.4\std{1.3} 
& \cellcolor{second} 40.5\std{0.7} & \cellcolor{second} 71.1\std{0.6} 
& \cellcolor{second} 30.2\std{0.3} & \cellcolor{best} \textbf{60.6}\std{0.5} 
& \cellcolor{second} 2.9 \\

\textsc{MedFeat} 
& \cellcolor{best} \textbf{11.7}\std{1.3} & \cellcolor{best} \textbf{77.9}\std{1.7} 
& \cellcolor{best} \textbf{26.4}\std{1.5} & \cellcolor{best} \textbf{70.2}\std{2.5} 
& \cellcolor{best} \textbf{13.6}\std{1.2} & \cellcolor{best} \textbf{68.8}\std{0.7} 
& \cellcolor{second} 1.8\std{0.3} & \cellcolor{best} \textbf{75.8}\std{2.7} 
& \cellcolor{best} \textbf{58.2}\std{0.6} & \cellcolor{best} \textbf{75.4}\std{1.1} 
& \cellcolor{best} \textbf{40.7}\std{0.6} & \cellcolor{best} \textbf{71.3}\std{0.5} 
& \cellcolor{best} \textbf{30.3}\std{0.6} & 59.4\std{1.9} 
& \cellcolor{best}{\textbf{1.2}} \\ 

\textit{Improv.\%} 
& \textcolor{colorgreen}{23.2\%} & \textcolor{colorgreen}{3.0\%}
&  \textcolor{colorgreen}{11.9\%} & \textcolor{colorgreen}{2.8\%} 
&  \textcolor{colorgreen}{5.4\%} &  \textcolor{colorgreen}{2.1\%} 
& \textcolor{colorgreen}{28.6\%} &  \textcolor{colorgreen}{6.8\%} 
&  \textcolor{colorgreen}{5.2\%} &  \textcolor{colorgreen}{2.2\%} 
& \textcolor{colorgreen}{2.3\%} &  \textcolor{colorgreen}{0.4\%}  
&  \textcolor{colorgreen}{2.0\%} & \textcolor{colorgreen}{1.2\%} 
&  \textcolor{colorgreen}{-2.9}\\

\midrule

\multicolumn{16}{@{}l}{\textit{Logistic Reg.}} \\

Baseline 
& 8.3\std{0.3} & 75.6\std{0.5} 
& 28.1\std{4.0} & \cellcolor{second} 75.0\std{1.9} 
& 12.2\std{0.3} & 66.3\std{0.8} 
& 2.2\std{0.3} & \cellcolor{second} 83.9\std{1.8} 
& 57.3\std{1.1} & \cellcolor{second} 75.6\std{1.0} 
& 38.4\std{0.5} & \cellcolor{second} 69.5\std{0.2} 
& 28.6\std{0.3} & 57.6\std{0.5} 
& 3.2 \\

OpenFE 
& 5.6\std{1.2} & 57.7\std{6.3} 
& 19.7\std{4.4} & 62.6\std{2.5} 
& 7.9\std{3.3} & 55.6\std{6.0} 
& \cellcolor{best} \textbf{3.8}\std{4.3} & 49.1\std{4.6} 
& 29.9\std{12.5} & 70.9\std{2.1} 
& 28.2\std{10.3} & 65.4\std{1.8} 
& 25.1\std{3.4} & 56.3\std{1.0} 
& 4.7 \\

CAAFE 
& 8.5\std{0.5} & 76.2\std{1.3} 
& \cellcolor{second} 28.2\std{4.1} & \cellcolor{second} 75.0\std{1.9} 
& 12.6\std{0.3} & 66.9\std{0.9} 
& 2.2\std{0.2} & \cellcolor{second} 83.9\std{1.8} 
& \cellcolor{second} 57.5\std{1.3} & \cellcolor{second} 75.6\std{1.0} 
& 38.4\std{0.5} & \cellcolor{second} 69.5\std{0.2} 
& 28.7\std{0.4} & 57.8\std{0.6} 
& \cellcolor{second} 2.9 \\

FeatLLM 
& \small{N/A} & 50.0\std{0.0} 
& \small{N/A} & 54.9\std{5.2} 
& \small{N/A} & 50.0\std{0.0} 
& \small{N/A} & 62.2\std{4.7} 
& \small{N/A} & 56.0\std{4.2} 
& \small{N/A} & 50.4\std{0.6} 
& \small{N/A} & 50.0\std{0.0} 
& 5.9 \\

OCTree 
& \cellcolor{second} 9.3\std{1.4} & \cellcolor{second} 77.1\std{1.0} 
& \cellcolor{second} 28.2\std{4.1} & \cellcolor{second} 75.0\std{1.9} 
& \cellcolor{second} 12.7\std{0.7} & \cellcolor{second} 67.0\std{1.1} 
& 2.2\std{0.2} & \cellcolor{second} 83.9\std{1.8} 
& \cellcolor{second} 57.5\std{0.9} & \cellcolor{second} 75.6\std{1.0} 
& \cellcolor{second} 38.5\std{0.5} & \cellcolor{second} 69.5\std{0.2} 
& \cellcolor{second} 28.9\std{0.3} & \cellcolor{second} 58.0\std{0.5} 
& 3.1 \\

\textsc{MedFeat} 
& \cellcolor{best} \textbf{10.9}\std{1.7} & \cellcolor{best} \textbf{78.6}\std{1.3} 
& \cellcolor{best} \textbf{31.3}\std{3.3} & \cellcolor{best} \textbf{75.5}\std{1.9} 
& \cellcolor{best} \textbf{12.9}\std{0.2} & \cellcolor{best} \textbf{67.9}\std{1.6} 
& \cellcolor{second} 2.6\std{0.8} & \cellcolor{best} \textbf{84.2}\std{1.7} 
& \cellcolor{best} \textbf{59.6}\std{1.1} & \cellcolor{best} \textbf{77.0}\std{1.0} 
& \cellcolor{best} \textbf{40.5}\std{0.8} & \cellcolor{best} \textbf{71.4}\std{0.2} 
& \cellcolor{best} \textbf{29.8}\std{0.1} & \cellcolor{best} \textbf{59.8}\std{0.6} 
& \cellcolor{best} {\textbf{1.1}} \\

\textit{Improv.\%} 
& \textcolor{colorgreen}{31.3\%} &  \textcolor{colorgreen}{4.0\%} 
& \textcolor{colorgreen}{11.4\%} &  \textcolor{colorgreen}{0.7\%} 
& \textcolor{colorgreen}{5.7\%} &  \textcolor{colorgreen}{2.4\%} 
& \textcolor{colorgreen}{18.2\%} & \textcolor{colorgreen}{0.4\%} 
& \textcolor{colorgreen}{4.0\%} &  \textcolor{colorgreen}{1.9\%} 
& \textcolor{colorgreen}{5.5\%} &  \textcolor{colorgreen}{2.7\%}  
& \textcolor{colorgreen}{4.2\%} &  \textcolor{colorgreen}{3.8\%} 
&  \textcolor{colorgreen}{-2.1} \\

\bottomrule
\end{tabular}}
\caption{Performance improvement by MedFeat across 7 clinical prediction tasks. \colorbox{best}{\textbf{Bold}} indicates best and \colorbox{second}{shaded} indicates second-best method. Avg. Rank computed across AUROC, F1, and AUPRC. MedFeat achieves the best performance under both classifiers significantly. \textit{Improv.} indicates improvement of MedFeat vs. the raw baseline. Mean and standard deviation
over 5 random splits.}
\label{tab:main_results}
\end{table*}

\subsection{Model-Aware Feature Generation}
For each island, MedFeat constructs a model-aware prompt that conditions feature generation on the downstream hypothesis class $\mathcal{H}$. Given the current downstream model $f_{\theta_t} \in \mathcal{H}$, MedFeat queries the LLM for the learner's representational properties and translates them into two complementary lists in the prompt: transformations to \emph{encourage} (those that expand $\mathcal{H}$'s effective capacity) and transformations to \emph{avoid} (those the learner can already recover from its own structure).

For example, for logistic regression, whose decision function is linear in the input, MedFeat encourages nonlinear transformations, interaction terms, and composite features that supply representational capacity the model cannot learn on its own, while discouraging trivial rescalings and monotonic remappings that leave the linear decision boundary unchanged. This learner-specific encourage/avoid structure steers proposals toward transformations $\sigma : \mathbb{R}^{M+t-1} \rightarrow \mathbb{R}$ with high marginal benefit for $f_{\theta_t}$.

\subsection{Candidate Evaluation and Selection}
Each candidate transformation generated from an island is executed and applied to $\mathcal{D}_{\text{train}}$ and $\mathcal{D}_{\text{val}}$. For valid candidates, a downstream model is trained on the augmented training data and evaluated.

Let $\mathcal{L}_{t,k}$ denote the validation score obtained from the $k$-th island at iteration~$t$, i.e.,
\[
  \mathcal{L}_{t,k}
  = L\!\bigl(f^{\star}_{\boldsymbol{\sigma}_t \cup \{\sigma_{t,k}\}},\;
             \mathcal{D}_{\text{val}} \oplus (\boldsymbol{\sigma}_t \cup \{\sigma_{t,k}\})\bigr).
\]
MedFeat selects the best-performing island as
$k^{\star} = \arg\max_{k}\, \mathcal{L}_{t,k}$.
If $\mathcal{L}_{t,k^{\star}} \geq \mathcal{L}_{t,\text{baseline}} - \beta$,
where $\beta \geq 0$ is a tolerance parameter that accounts for noise in the validation set,
the feature transformations generated by island~$k^{\star}$ are accepted and merged into~$\boldsymbol{\sigma}_{t+1}$.
Otherwise, all candidate transformations from the iteration are rejected.
Upon acceptance, the baseline model is retrained on the updated feature set, and new feature relevance scores are computed, closing the feedback loop.

\subsection{Memory and Termination}

MedFeat maintains a memory of accepted and rejected feature transformations across iterations.
Rejected features are tracked to discourage repeated low-value proposals, while accepted features
influence future relevance estimates and island sampling. This improves search behavior by reinforcing feature patterns that consistently enhance performance or importance and increasing computational efficiency by reducing redundant exploration \cite{nam2024optimized}.
The procedure terminates after $T$ iterations, yielding a final augmented model trained on
$\mathcal{D}_{\text{train}} \oplus \boldsymbol{\sigma}_T$, which is then evaluated once on the
held-out test set.

\begin{table*}[]
\centering
\renewcommand{\arraystretch}{1}
\resizebox{\textwidth}{!}{%
\begin{tabular}{@{}c*{7}{cc}c@{}}
\toprule
\multirow{2}{*}[-0.8ex]{\textbf{Model}}
& \multicolumn{3}{c}{\textbf{F1}}
& \multicolumn{3}{c}{\textbf{AUROC}}
& \multicolumn{3}{c}{\textbf{AUPRC}} \\
\cmidrule(lr){2-4} \cmidrule(lr){5-7} \cmidrule(lr){8-10}
  & {MedFeat} & {Baseline} & {$\Delta$\% [min, max]}
  & {MedFeat} & {Baseline} & {$\Delta$\% [min, max]}
  & {MedFeat} & {Baseline} & {$\Delta$\% [min, max]} \\
\midrule
Logistic Reg.
  & 26.6{\scriptsize$\pm$1.1} & 25.8{\scriptsize$\pm$1.1} & {$5.5\ [1.5,\ 20.7]$}
  & 73.0{\scriptsize$\pm$1.3} & 72.2{\scriptsize$\pm$1.2} & {$1.1\ [0.1,\ 2.3]$}
  & 23.5{\scriptsize$\pm$0.9} & 22.6{\scriptsize$\pm$0.9} & {$4.8\ [1.8,\ 19.0]$} \\
XGBoost
  & 27.8{\scriptsize$\pm$1.1} & 27.3{\scriptsize$\pm$1.0} & {$3.1\ [0.4,\ 8.7]$}
  & 75.3{\scriptsize$\pm$1.4} & 74.8{\scriptsize$\pm$1.4} & {$0.6\ [0.2,\ 1.8]$}
  & 27.5{\scriptsize$\pm$1.8} & 27.0{\scriptsize$\pm$1.7} & {$3.8\ [0.5,\ 16.2]$} \\
\midrule
\rowcolor{overallgray}
\textbf{Overall}
  & \textbf{27.2}{\scriptsize$\pm$1.1} & \textbf{26.6}{\scriptsize$\pm$1.0} & {$\mathbf{4.3\ [0.4,\ 20.7]}$}
  & \textbf{74.1}{\scriptsize$\pm$1.4} & \textbf{73.5}{\scriptsize$\pm$1.3} & {$\mathbf{0.8\ [0.1,\ 2.3]}$}
  & \textbf{25.5}{\scriptsize$\pm$1.4} & \textbf{24.8}{\scriptsize$\pm$1.3} & {$\mathbf{4.3\ [0.5,\ 19.0]}$} \\
\bottomrule
\end{tabular}%
}
\caption{Performance summary across models and metrics(\%) \textbf{after extensive HPO}. Metrics are averaged over all 7 tasks and 2 models. $\Delta$\% = mean relative improvement; brackets show [min, max] range across tasks. Mean and standard deviation
over 5 random splits.}
\label{tab:hpo_summary}
\end{table*}

\section{Experiments}
\subsection{Datasets and Tasks}
We benchmark on three well-established healthcare datasets spanning both hospital electronic health records and longitudinal panel data: the Medical Information Mart for Intensive Care (MIMIC) IV, the Infections in Oxfordshire Research Database (IORD), and the Health and Retirement Study (HRS) \cite{wei2024predicting, johnson2023mimic, PhysioNet-mimiciv-3.1, goldberger2000physiobank, hrs_dataset}. Each dataset is preprocessed following established pipelines from prior studies \cite{puterman2020predicting, gupta2022extensive}. Our datasets also provide a wide coverage of clinical outcomes, including short and long-term mortality, phenotypes, discharge, and readmission prediction. A key motivation for this suite is that the feature space is heterogeneous: the datasets include both static covariates and repeated measurements at different time points with irregular missingness patterns. Details of the datasets and tasks are provided in Appendix~\ref{app:data}.

\subsection{Baselines and Downstream Models}
We compare MedFeat against a hierarchy of baselines designed to isolate the contribution of (i) feature engineering in general, (ii) classical automated feature engineering, and (iii) recent LLM-based feature engineering methods. Concretely, we compare models trained on the raw feature set; models with classical automated feature engineering baselines (e.g., AutoFeat \cite{horn2019autofeat}, OpenFE \cite{zhang2023openfe}); and models with LLM-based baselines (CAAFE \cite{hollmann2023large}, OCTree \cite{nam2024optimized}, FeatLLM \cite{han2024large}), alongside MedFeat. We evaluate the performance in two settings: a setting with the model's default hyperparameters, and a setting in which extensive hyperparameter optimization (HPO) is applied to quantify the best achievable performance and assess whether feature engineering gains persist under HPO. 

We focus on two widely used tabular learners with complementary inductive biases: logistic regression (LR) \cite{christodoulou2019systematic} and XGBoost (XGB). Logistic regression is widely used in clinical research and is valued for its explainability and lightweight nature, and sometimes offers superior performance to complex machine learning models \cite{christodoulou2019systematic}. XGBoost is a powerful tree-based model that achieves highly competitive performance against neural methods on clinical tabular data \cite{chen2016xgboost, grinsztajn2022tree}.

\subsection{Implementation}
We use a stratified random split with 60\% of the data as the training set, 20\% as the validation set, and 20\% as the held-out test set. The validation set is used both for selecting successful feature transformations and for computing explainability signals. We evaluate on metrics such as Area Under the Receiver Operating Characteristic curve (AUROC), F1 score, and Area Under the Precision-Recall Curve (AUPRC) to account for class imbalance and to capture both ranking ability and precision-recall trade-offs at the decision boundary \cite{davis2006relationship}, since rare outcomes remain a central difficulty when modeling healthcare datasets. Further implementation details are provided in Appendix~\ref{app:implementations}.

\section{Main Results}

Table \ref{tab:main_results} presents the main results across seven clinical prediction tasks evaluated under two downstream classifiers. MedFeat consistently outperforms all baselines and yields substantial performance gains across tasks. Specifically, the F1 score achieves an average improvement of 11.4\% over no feature engineering. MedFeat attains the best average rank under both logistic regression (1.1) and XGBoost (1.2). MedFeat has demonstrated statistically significantly better performance than all competing methods (Details and statistical tables in Appendix \ref{app:significance}). Our key observations are organized below:

\paragraph{MedFeat yields the largest gains on the most difficult tasks} MedFeat demonstrates the largest F1 improvements on the most imbalanced tasks with rare positive outcomes: Mortality-ICU (31.3\% under LR, 23.2\% under XGB), Mortality-Ward (18.2\%, 28.6\%), and 10-year Mortality (11.4\%, 11.9\%). Combined with the observation that performance gains are more pronounced on F1 than on AUROC, this pattern suggests that the engineered features are particularly effective at surfacing predictive signals for rare positive cases. For tasks such as Cognitive Impairment and Discharge, existing approaches often fail to improve the logistic regression baseline, or even introduce noise that degrades performance relative to the raw baseline; in contrast, MedFeat still achieves consistent gains (1.9\% for Cognitive Impairment and 2.7\% for Discharge). These gains are valuable in settings where competing methods fail to perform, even though their absolute magnitude is relatively modest. The only exception is F1 on Mortality-Ward, where OpenFE reports a higher mean (6.1). However, its standard deviation (±4.1) is nearly as large as the mean itself, suggesting that the result may be driven by a few unstable runs.

\paragraph{MedFeat demonstrates better stability across runs} MedFeat consistently delivers superior performance across tasks. Among the baselines, OCTree and CAAFE alternately occupy the second-best position, yet their average ranks (2.9–3.2) trail MedFeat by a substantial margin, and the only exception is OCTree achieving a slightly better AUROC on Readmission under XGBoost. FeatLLM fails on multiple tasks, and OpenFE exhibits large performance fluctuations across tasks, frequently underperforming simple baselines despite rarely achieving competitive performance. Beyond higher mean performance, MedFeat’s standard deviations are almost always comparable to or smaller than those of competitive baselines. For example, on 10-year Mortality under logistic regression, MedFeat reports an 11.4\% improvement over the baseline while having the smallest standard deviation (3.3). Similar patterns hold for tasks such as Cognitive Impairment. This is particularly valuable given that severe class imbalance and noise naturally amplify variance across runs, yet MedFeat maintains tighter confidence intervals than methods operating under the same conditions.

\paragraph{MedFeat expands the capabilities of weaker learners.} In many cases, the performance gains from MedFeat are larger under logistic regression than under XGBoost. For instance, Mortality-ICU F1 improves by 31.3\% under LR versus 23.2\% under XGBoost, and Discharge AUROC improves by 2.7\% versus 0.4\%. Furthermore, on tasks such as Readmission, logistic regression augmented by MedFeat (AUROC 57.6 → 59.8) surpasses the XGBoost baseline (AUROC 57.6). A similar pattern emerges on Mortality-ICU, where logistic regression’s AUROC rises from 75.6 to 78.6, and XGBoost’s rises from 75.6 to 77.9. This indicates that the model-aware features compensate more effectively when the downstream learner has limited representational capacity, supplying nonlinear and interaction-based signals that a linear model cannot recover on its own.

\section{Additional Analysis}

\paragraph{Feature engineering gains persist under hyperparameter optimization} MedFeat maintains consistent gains across all metrics and both classifiers: overall F1 improves by 4.3\%, AUROC by 0.8\%, and AUPRC by 4.3\% (Table \ref{tab:hpo_summary}). Notably, the minimum improvement across all tasks–model–metric combinations remains positive, confirming that no task regresses after HPO. The larger F1 and AUPRC gains (4.3\%) reveal that MedFeat's features continue to benefit minority-class precision after HPO. The wide upper bounds of the improvement ranges (up to 20.7\% for F1) further indicate that MedFeat can contribute substantial gains on some tasks even after HPO. Since HPO directly maximizes AUROC, the residual 0.8\% AUROC improvement represents a signal that tuning alone cannot recover from the original feature space, where the engineered features expand the representational capacity available to the learner. 

\begin{table}[ht]
\centering
\small
\renewcommand{\arraystretch}{1.15}
\resizebox{\columnwidth}{!}{%
\begin{tabular}{cccc}
\toprule
\textbf{Ablation Mode} & \textbf{F1} ($\Delta$\,\%) & \textbf{AUROC} ($\Delta$\,\%) & \textbf{AUPRC} ($\Delta$\,\%) \\
\midrule
w/o SHAP Importance       & 25.3 {\scriptsize$\pm$1.0} (--4.5\%) & 71.3 {\scriptsize$\pm$1.2} (--1.5\%) & 22.8 {\scriptsize$\pm$0.9} (--4.2\%) \\
w/o Model-Awareness      & 25.6 {\scriptsize$\pm$1.0} (--3.4\%) & 71.0 {\scriptsize$\pm$1.2} (--1.9\%) & 23.2 {\scriptsize$\pm$1.0} (--2.5\%) \\
w/o Island Sampling          & 25.8 {\scriptsize$\pm$1.1} (--2.6\%) & 71.1 {\scriptsize$\pm$1.3} (--1.8\%) & 23.3 {\scriptsize$\pm$1.0} (--2.1\%) \\
w/o Memory           & 25.8 {\scriptsize$\pm$1.0} (--2.6\%) & 71.1 {\scriptsize$\pm$1.2} (--1.8\%) & 23.3 {\scriptsize$\pm$1.0} (--2.1\%) \\
\bottomrule
\end{tabular}%
}
\caption{Ablation study summary. Values in parentheses denote performance drop (\%) relative to the full model. Mean and standard deviation
over 5 random splits.}
\label{tab:ablation}
\end{table}

\paragraph{Ablation study confirms the contribution of each component} Table \ref{tab:ablation} reports the effect of removing each core module, averaged across all tasks and both classifiers. Every ablation leads to a consistent decline across all three metrics, confirming that each component contributes positively, and no removal fully erases the gains reported in Table \ref{tab:main_results}. SHAP-based importance produces the largest F1 and AUPRC drops (-4.5\%, -4.2\%), consistent with our finding that MedFeat’s gains concentrate on threshold-dependent metrics sensitive to rare outcomes when the framework focuses on importance-weighted features. Model-awareness is the only ablation where the AUROC drop (-1.9\%) exceeds that of the SHAP ablation, suggesting that model-awareness and SHAP importance affect partially different aspects of predictive quality. Island sampling and feedback memory each produce smaller, comparable drops, indicating similar contributions. The standard deviations also increase slightly in most ablated configurations relative to the full model, suggesting that each removed component contributes not only to performance but also to the stability of the feature engineering process.

\paragraph{MedFeat generalizes across different LLM backbones} Table \ref{tab:cross_llm_comparison} compares MedFeat against CAAFE, OCTree, and FeatLLM when varying the underlying language model across four representative backbones spanning proprietary and open-weight families. MedFeat achieves the highest performance under every backbone tested, confirming that its gains are from the framework design rather than from a particular LLM's capabilities. Importantly, MedFeat paired with the smallest backbone, Qwen-3.6-27B, still outperforms all baselines running on substantially larger proprietary models such as GPT-4o and GPT-5.4 on every metric (71.5 vs.\ 71.0 AUROC; 23.9 vs.\ 22.5 AUPRC against OCTree-GPT-4o), demonstrating that the structured feedback loop effectively compensates for reduced LLM capacity and lowers the deployment cost in privacy-sensitive or on-premise clinical settings where open-weight models are required. We also observe that GPT-5.4 yields MedFeat's strongest AUPRC (24.8), consistent with its stronger compositional reasoning. The performance gap between backbones within MedFeat is consistently narrow: AUROC ranges only from 71.5 to 72.4 (a spread of 0.9 points), F1 from 25.9 to 26.5, and AUPRC from 23.6 to 24.8. Competing methods exhibit similarly tight inter-backbone variation but have a lower upper bound, indicating that backbone scale alone cannot close the gap left by designs. The improvement of MedFeat over the next-best method remains stable across backbones. FeatLLM continues to underperform substantially regardless of backbone, suggesting that its rule-based binary feature design, rather than backbone choice, is the limiting factor on imbalanced clinical tasks.

\begin{table*}[h]
\centering
\small
\resizebox{\textwidth}{!}{%
\begin{tabular}{c ccc ccc cc ccc}
\toprule
& \multicolumn{3}{c}{CAAFE} & \multicolumn{3}{c}{OCTree} & \multicolumn{2}{c}{FeatLLM} & \multicolumn{3}{c}{\textbf{MedFeat (Ours)}} \\
\cmidrule(lr){2-4} \cmidrule(lr){5-7} \cmidrule(lr){8-9} \cmidrule(lr){10-12}
\multirow{-2}{*}{\centering\textbf{LLM Backbone}} & F1 & AUROC & AUPRC & F1 & AUROC & AUPRC & AUROC & AUPRC & F1 & AUROC & AUPRC \\
\midrule
gpt-4o-2024-11-20
  & 25.1 \scriptsize{$\pm$0.9} & 70.8 \scriptsize{$\pm$1.3} & 22.5 \scriptsize{$\pm$1.1}
  & 25.4 \scriptsize{$\pm$1.0} & 71.0 \scriptsize{$\pm$1.2} & 22.5 \scriptsize{$\pm$0.9}
  & 52.2 \scriptsize{$\pm$1.6} & 12.2 \scriptsize{$\pm$0.7}
  & \textbf{26.5} \scriptsize{$\pm$\textbf{1.0}} & \textbf{72.4} \scriptsize{$\pm$\textbf{1.4}} & \textbf{23.8} \scriptsize{$\pm$\textbf{0.9}} \\
gpt-5.4
  & 25.2 \scriptsize{$\pm$1.0} & 70.7 \scriptsize{$\pm$1.2} & 22.5 \scriptsize{$\pm$1.0}
  & 25.8 \scriptsize{$\pm$1.0} & 70.6 \scriptsize{$\pm$0.9} & 21.8 \scriptsize{$\pm$0.8}
  & 53.8 \scriptsize{$\pm$2.8} & 13.3 \scriptsize{$\pm$1.4}
  & \textbf{26.5} \scriptsize{$\pm$\textbf{0.9}} & \textbf{72.2} \scriptsize{$\pm$\textbf{1.2}} & \textbf{24.8} \scriptsize{$\pm$\textbf{1.3}} \\
gemini-3.1-pro-preview
  & 25.2 \scriptsize{$\pm$0.8} & 70.8 \scriptsize{$\pm$1.3} & 22.5 \scriptsize{$\pm$0.9}
  & 25.8 \scriptsize{$\pm$1.0} & 70.8 \scriptsize{$\pm$1.1} & 21.9 \scriptsize{$\pm$0.8}
  & 53.5 \scriptsize{$\pm$2.3} & 13.0 \scriptsize{$\pm$0.9}
  & \textbf{25.9} \scriptsize{$\pm$\textbf{1.0}} & \textbf{72.1} \scriptsize{$\pm$\textbf{1.2}} & \textbf{23.6} \scriptsize{$\pm$\textbf{1.0}} \\
qwen-3.6-27B-FP8
  & 25.0 \scriptsize{$\pm$0.9} & 70.5 \scriptsize{$\pm$1.1} & 22.4 \scriptsize{$\pm$0.9}
  & 25.8 \scriptsize{$\pm$1.0} & 70.7 \scriptsize{$\pm$0.9} & 21.8 \scriptsize{$\pm$0.8}
  & 52.2 \scriptsize{$\pm$2.2} & 12.3 \scriptsize{$\pm$1.0}
  & \textbf{26.2} \scriptsize{$\pm$\textbf{1.0}} & \textbf{71.5} \scriptsize{$\pm$\textbf{1.1}} & \textbf{23.9} \scriptsize{$\pm$\textbf{1.0}} \\
\bottomrule
\end{tabular}
}
\caption{Performance comparison of \textbf{MedFeat} against CAAFE, OCTree, and FeatLLM across different LLM backbones. Results are reported as mean and std over 7 tasks and 2 models. Best results per LLM and metric are shown in \textbf{bold}. FeatLLM does not report F1.}
\label{tab:cross_llm_comparison}
\end{table*}

\begin{figure}[]
  \centering
  \includegraphics[width=1\columnwidth]{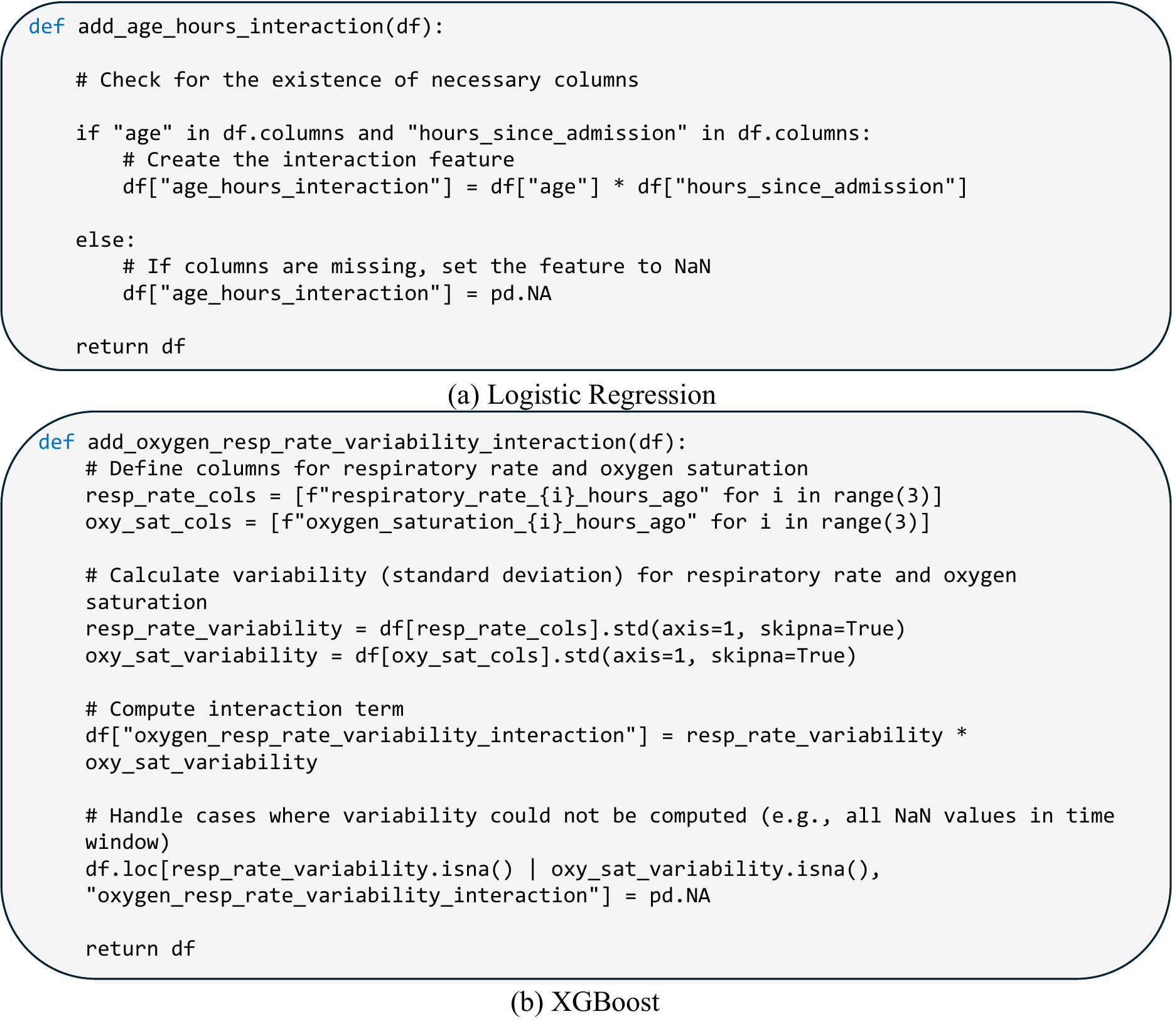}
  \caption{Feature transformation code generated by MedFeat for logistic regression and XGBoost}
  \label{fig:code}
\end{figure}

\paragraph{Clinical evidence and model awareness} To illustrate the clinical relevance and model-awareness of MedFeat-generated features, we examine two transformations proposed for the same discharge prediction task under different learners (Figure \ref{fig:code}). For logistic regression, MedFeat generates an \textit{age} and \textit{length of stay in hospital} interaction, capturing the clinically established phenomenon that older hospitalized patients have a higher risk \cite{elgar2022major}, which is a nonlinear relationship that a linear model cannot recover from the raw covariates alone. For XGBoost, MedFeat instead proposes a \textit{respiratory rate} and \textit{oxygen saturation} variability interaction, reflecting the physiological coupling between compensatory breathing effort and deteriorating oxygenation, supported by clinical literature \cite{garrido2018respiratory, rivas2023early, bhogal2017pattern}, which is a high-order temporal pattern that trees struggle to learn from individual columns. These two features target the same clinical decision yet differ in construction: this contrast exemplifies how model-aware prompting guides MedFeat toward transformations that are both clinically grounded and structurally complementary to the downstream model.

\begin{figure}[h]
  \centering
  \hspace{-0.05\columnwidth}
  \includegraphics[width=1\columnwidth]{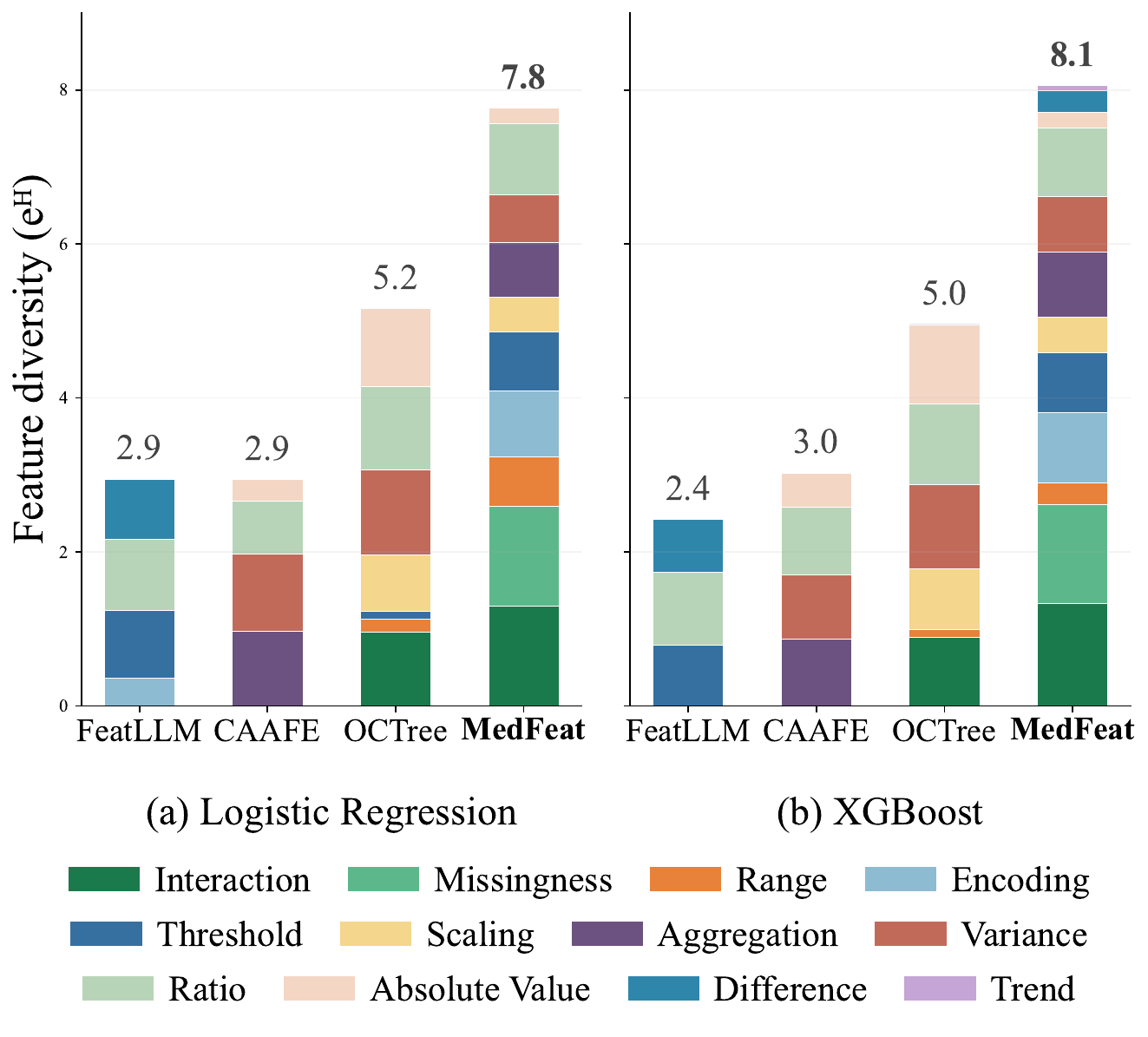}
  \caption{Feature patterns of the LLM-based feature engineering generated features for two models}
  \label{fig:feature_diversity}
\end{figure}

\paragraph{MedFeat generates diversified transformation patterns} In Figure \ref{fig:feature_diversity}, we categorize all accepted features into 12 pattern types and quantify diversity using the exponential Shannon entropy $e^H$ \cite{Entropy2004ShannonER, renyi1961measures}. MedFeat achieves the highest feature diversity under both classifiers (7.8 for LR, 8.1 for XGBoost), substantially exceeding all other methods. Baseline methods concentrate on narrow pattern subsets: FeatLLM relies heavily on clinical thresholds and interactions, while CAAFE allocates over half its outputs to aggregation and ranging. All baseline methods produce nearly identical distributions across both classifiers, confirming learner-agnostic design. In contrast, MedFeat distributes proposals across all 12 categories and adapts its pattern distribution to each learner’s inductive bias. Under logistic regression, ranging transformation constitutes 19.2\% of accepted blocks but drops to 6.1\% under XGBoost, since compressing temporal spread into a scalar introduces nonlinear patterns, whereas trees approximate such patterns through successive splits. Conversely, aggregation rises from 22.1\% to 31.3\% under XGBoost, and trend features emerge only under XGBoost, supplying cross-temporal summaries and sequential dynamics. These shifts confirm that model-aware prompting diversifies the feature types and guides MedFeat toward structurally complementary transformations.

\begin{figure}[h]
  \centering
  \includegraphics[width=0.8\columnwidth]{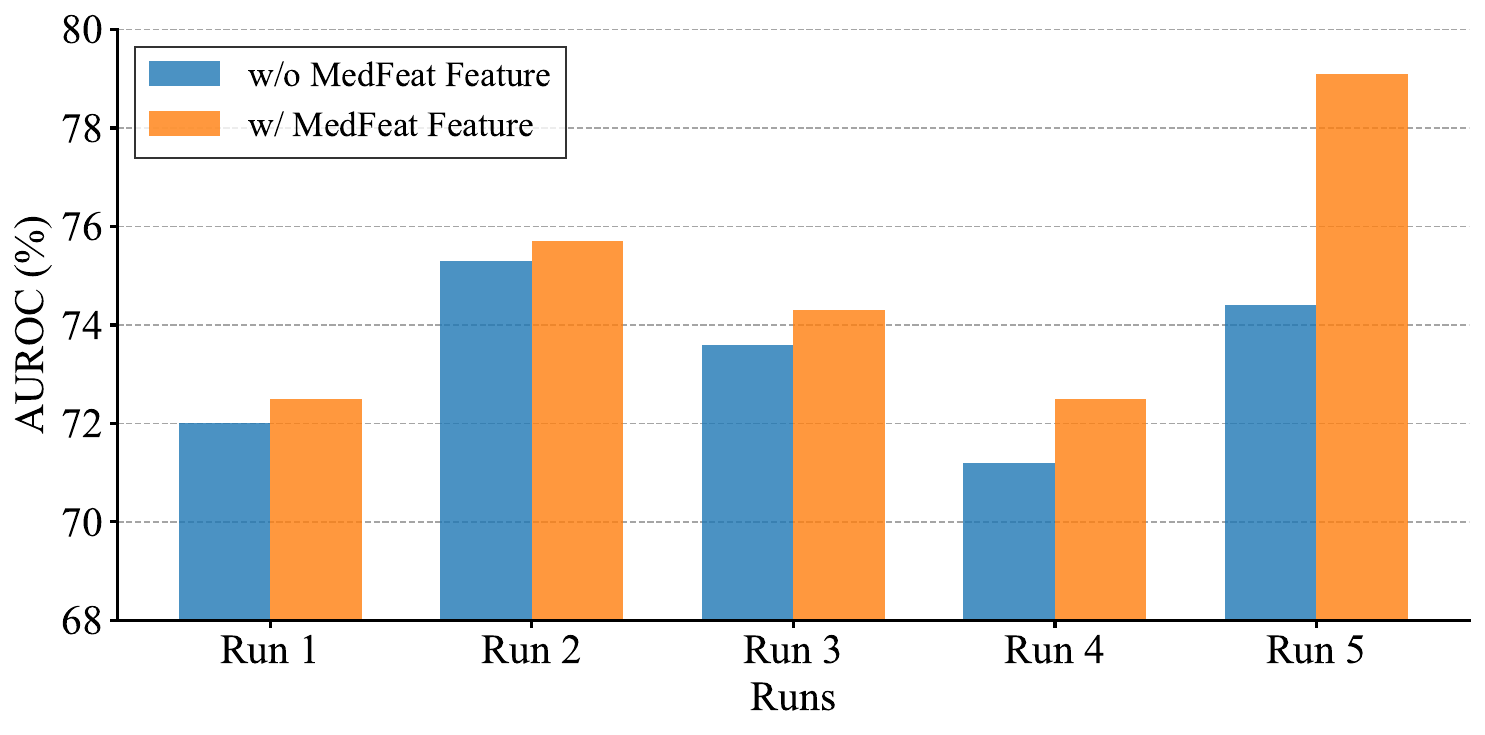}
  \caption{Evaluations on IORD data enhanced by transferring MedFeat-generated features from MIMIC}
  \label{fig:transfer}
\end{figure}

\paragraph{MedFeat-generated features transfer across clinical cohorts} Figure \ref{fig:transfer} reports AUROC across five independent runs when MedFeat features transformation generated on MIMIC-IV are applied to an XGBoost model on IORD. Transferred features include trends, ratios, and demographic interactions. All five runs improve, with gains of 0.4–4.7 AUROC (largest on Run 5, 74.4 → 79.1). The consistent direction of improvement despite substantial baseline variation across splits indicates the gains reflect the transferred features rather than split-specific noise, suggesting MedFeat captures clinically meaningful signals shared across ICU and general hospitalized populations rather than overfitting to MIMIC. From a deployment perspective, this transferability lowers the barrier to adoption in resource-constrained settings: institutions without access to LLM infrastructure or sufficient patient volume can directly inherit validated features from larger reference cohorts, without sharing patient data.

\section{Conclusion}

MedFeat reframes LLM-driven feature engineering as a model-aware, attribution-guided search, replacing the learner-agnostic design of existing methods with an iterative loop that mirrors practitioners' practice: train, interpret, localize, propose, evaluate, and reflect. Extensive experiments across a diverse range of clinical prediction tasks demonstrate that MedFeat achieves statistically significant improvements over all existing methods, with particularly pronounced gains on severely imbalanced tasks and weaker learners. Crucially, gains persist under heavy hyperparameter tuning, reflect diversified transformation patterns, transfer across cohorts without sharing patient data, and generalize across multiple LLM backbones. By unifying domain knowledge, model feedback, and attribution signals into a principled loop, MedFeat offers a novel and practical paradigm for automated clinical feature engineering while remaining compatible with privacy-constrained deployment.

\section*{Limitations}

One potential limitation of MedFeat is that the framework relies on explanation-based signals to guide sampling and prompting. If the data has strong collinearity or noise, the importance ordering might vary across splits. This can be mitigated by combining the framework with pre-hoc classical feature selection methods such as L1 and L2 regularization, which can be easily integrated with MedFeat pipeline. Additionally, model-awareness in MedFeat is implemented as
learner-conditioned prompting over the hypothesis class;
diagnosing the specific representational gaps of a trained
model is a natural
extension we leave to future work.

\bibliography{custom}

\clearpage

\appendix

\section{Algorithm}
\label{app:algorithm}

We provide the full algorithm of MedFeat below. The algorithm summarizes the iterative explainability-guided island sampling, LLM-based feature proposal, and local evaluation procedures used throughout our experiments. 

\begin{algorithm}[H]
\caption{\textsc{MedFeat}}
\label{alg:MedFeat}

\begin{algorithmic}

\STATE {\bfseries Input:} $\mathcal{D}_{\text{train}}, \mathcal{D}_{\text{val}}$, hypothesis class $\mathcal{H}$,
metric $\mathcal{L}$, explainability function $\mathrm{Expl}$, LLM $\mathcal{M}$,
iterations $T$, islands $K$, island size $m$, tolerance $\beta$.
\STATE {\bfseries Output:} $\boldsymbol{\sigma}_T, f^\star$

\STATE $\boldsymbol{\sigma}_0 \leftarrow \emptyset$
\STATE $f^\star \leftarrow \arg\max_{f\in\mathcal{H}} \mathcal{L}(f,\mathcal{D}_{\text{train}})$
\STATE $\mathcal{L}_{\text{base}} \leftarrow \mathcal{L}(f^\star,\mathcal{D}_{\text{val}})$
\STATE $s \leftarrow \mathrm{Expl}(f^\star,\mathcal{D}_{\text{val}})$

\FOR{$t=1$ {\bfseries to} $T$}
  \FOR{$k=1$ {\bfseries to} $K$}
    \STATE $I_{t,k} \sim \text{Sample}(s, m)$
    \STATE $\sigma_{t,k} \leftarrow \mathcal{M}(I_{t,k})$
    \STATE $f_{t,k}^\star \leftarrow \arg\max_{f\in\mathcal{H}}
    \mathcal{L}(f,\mathcal{D}_{\text{train}}\oplus\boldsymbol{\sigma}_{t-1}\cup\sigma_{t,k})$
    \STATE $\mathcal{L}_{t,k} \leftarrow
    \mathcal{L}(f_{t,k}^\star,\mathcal{D}_{\text{val}}\oplus\boldsymbol{\sigma}_{t-1}\cup\sigma_{t,k})$
  \ENDFOR

  \STATE $k^\star \leftarrow \arg\max_k \mathcal{L}_{t,k}$
  \IF{$\mathcal{L}_{t,k^\star} \ge \mathcal{L}_{\text{base}} - \beta$}
    \STATE $\boldsymbol{\sigma}_t \leftarrow \boldsymbol{\sigma}_{t-1} \cup \sigma_{t,k^\star}$
    \STATE $f^\star \leftarrow f_{t,k^\star}^\star$
    \STATE $\mathcal{L}_{\text{base}} \leftarrow \mathcal{L}_{t,k^\star}$
    \STATE $s \leftarrow \mathrm{Expl}(f^\star,\mathcal{D}_{\text{val}})$
  \ELSE
    \STATE $\boldsymbol{\sigma}_t \leftarrow \boldsymbol{\sigma}_{t-1}$
  \ENDIF
\ENDFOR

\STATE {\bfseries Return} $\boldsymbol{\sigma}_T, f^\star$

\end{algorithmic}
\end{algorithm}

\section{Prompts}
Below, we provide an exemplar prompt template for MedFeat framework.

\begin{promptbox}[MedFeat Prompt Template Example]
You are an experienced clinical data scientist. Your task is to design new numerical features that complement the XGBoost and improve performance on the following classification task: predict whether the patient will die within 1 day.
\\
\\
Core principles you MUST follow:\\
1. Produce features that are difficult to learn for XGBoost, such as row-wise grouping and aggregation, global summary statistics over multiple features, complex temporal trending....\\
2. Avoid producing features that are easy to learn for XGBoost, such as simple thresholding, trivial interaction, and naive scaling.\\
3. Use the given SHAP-based feature importance to guide the feature engineering process. Prioritise features that are likely to improve performance.\\
4. Use evidence-based clinical reasoning to ground the feature.\\
5. Prioritise features similar to previous accepted features and avoid features similar to failed features.\\

Output requirements: Your answer must be a Python function snippet wrapped in ```python ... ``` and nothing else.
\\
\\
Dataset description:
...\\
Available features and their importance:
...\\
Previously successful features:
...\\
Previously failed features:
...\\
\end{promptbox}

\section{MedFeat Example Run}
Below, we provide an exemplar running log of the MedFeat framework. Numbers are illustrative.

\begin{promptbox}[MedFeat Agent Example]

{\small\itshape You are an experienced clinical data scientist\ldots}

\vspace{4pt}
\hrule
\vspace{4pt}

{\small\ttfamily Feature importance (SHAP values):}
\begin{itemize}[nosep, leftmargin=12pt, label={--}]
  \item \logline{age, importance: 0.1114, rank 1/34}
  \item \logline{index\_of\_multiple\_deprivation\_score, importance: 0.0837, rank 3/34}
  \item \logline{n\_unique\_meds\_received\_last\_24h, importance: 0.0274, rank 14/34}
\end{itemize}

\vspace{4pt}
\hrule
\vspace{4pt}

{\small\ttfamily $\rightarrow$ Attempt 1/3 -- generated code:}

\begin{lstlisting}[style=promptcode]
def add_age_imd_interaction_normalized(df):
    """
    Feature: Age * IMD Interaction (Normalized)
    Clinical Rationale: Age and deprivation
    compound frailty and limited access...
    """
    age_min = df['age'].min()
    age_max = df['age'].max()
    df['age_norm'] = (df['age'] - age_min)
                   / (age_max - age_min)
    ...
    return df
\end{lstlisting}

\vspace{4pt}
\hrule
\vspace{4pt}

{\small\ttfamily $\rightarrow$ Performing transformation and
training XGBoost model}\\
\logline{New Validation AUROC: 0.758,
  Baseline: 0.736
  $\rightarrow$ \textbf{age\_imd\_interaction kept}}

\vspace{6pt}
{\small\ttfamily Final Results:}
\begin{itemize}[nosep, leftmargin=12pt, label={--}]
  \item \logline{MedFeat AUROC: 0.707,
        Baseline AUROC: 0.669}
  \item \logline{MedFeat F1: 0.029,
        Baseline F1: 0.016}
\end{itemize}

\end{promptbox}

\section{Additional Implementation Details}
\label{app:implementations}

\paragraph{MedFeat configuration}

We use the following default configuration for the iterative search: 3 feature engineering iterations, 2 islands per iteration, and an island size of 3, with features sampled proportionally to normalized SHAP importance. We use TreeSHAP for XGBoost and a Linear explainer for the logistic regression model. Feature relevance is computed from mean absolute SHAP values. We use $\beta=0.01$ as the tolerance threshold by default.
The F1 score was computed using a validation-selected threshold (Youden’s $J$).
GPT-4o is the primary language model we use to generate feature transformations as executable Python functions for main results unless stated elsewhere \cite{hurst2024gpt}. Temperature was set to 0.7 where configurable.

\paragraph{Island generation} 
We treat repeated measurements of the same underlying variable as a temporal group when computing and averaging explainability signals for feature selection to account for collinearity. This approach addresses the collinearity of temporal features in clinical tabular data and promotes LLMs to discover subtle patterns in time series \cite{lundberg2017unified, liao2022does}. When creating each feature island, we randomly choose whether to sample static features or aggregated temporal groups. If a temporal group is chosen, all raw measurement features will be included in that island.

\paragraph{MedFeat transformation} 
For each candidate transformation, we execute the generated code in a restricted computing environment and apply it \emph{independently} to $\mathcal{D}_{\text{train}}$, $\mathcal{D}_{\text{val}}$, and $\mathcal{D}_{\text{test}}$ to prevent data leakage. We reject invalid transformations (non-numeric outputs, infinities/NaNs beyond handling rules, runtime errors) and allow up to three retries for transient generation failures. 

\paragraph{Downstream model common setup} Logistic regression is trained with an $\ell_2$ penalty, a Limited-memory BFGS solver, and a balanced class weight, reflecting a setting where linear models benefit from explicit feature construction but cannot recover non-linear structure on their own. Before fitting the logistic regression model, missing values in the original data were imputed using the median in the training set. Then, the data is standardized based on the scaler fitted on the training set. XGBoost uses the \texttt{scale\_pos\_weight} parameter, with the ratio being the ratio between negative and positive samples. XGBoost further employs histogram-based tree methods and binary logistic as objective functions. No imputation was performed for the XGBoost model unless stated elsewhere \cite{chen2016xgboost}.

\paragraph{Hyperparameter optimization}
We introduce comprehensive hyperparameter optimization to validate the effectiveness of the framework. We use Optuna with a budget of 400 trials and an early stopping mechanism. We use a random sampler and maximize AUROC. Besides the common setup mentioned above for downstream models, the search spaces following previous studies for XGBoost and logistic regression \cite{akiba2019optuna, grinsztajn2022tree, nam2024optimized} are presented in Table~\ref{tab:xgb_params},~\ref{tab:lr_params}.

\begin{table}[h]
\centering

\begin{tabular}{ll}
\toprule
\textbf{Parameter} & \textbf{Range [Min, Max]} \\
\midrule
max\_depth & [2, 7] \\
n\_estimators & [100, 2000] \\
min\_child\_weight & [10, 100] \\
max\_delta\_step & [0, 10] \\
subsample & [0.5, 1.0] \\
learning\_rate & [0.01, 0.5] \\
colsample\_bylevel & [0.5, 1.0] \\
colsample\_bytree & [0.3, 1.0] \\
gamma & [0, 5] \\
reg\_alpha & [0.5, 10.0] \\
reg\_lambda & [2, 20.0] \\
\bottomrule
\end{tabular}
\caption{Hyperparameter search space for XGBoost.}
\label{tab:xgb_params}
\end{table}

\begin{table}[h]
\centering

\begin{tabular}{ll}
\toprule
\textbf{Parameter} & \textbf{Range [Min, Max]} \\
\midrule
C & [$10^{-4}$, $10^{4}$] \\
\bottomrule
\end{tabular}
\caption{Hyperparameter search space for logistic regression.}
\label{tab:lr_params}
\end{table}

\paragraph{Baseline implementations}
In this section, we summarize the implementation and configuration of baseline feature engineering methods in our study. We implemented all baseline feature engineering methods by following their original implementations and default recommended settings and evaluated them under the same data splits, preprocessing, and downstream learners as MedFeat. For baselines that are iteration-based, we ran 3 iterations to match MedFeat’s feature-engineering rounds unless stated otherwise.

\paragraph{AutoFeat} We introduced AutoFeat as a classical automated feature engineering baseline. However, most of the execution runs exceeded 12 hours and failed to return results, leading to an insufficient set of results for a fair comparison. This runtime issue has also been noted in prior work when applying exhaustive operator-based feature engineering to wide real-world tables \cite{abhyankar2025llm}. Therefore, we excluded AutoFeat from our study due to its inapplicability.

\paragraph{OpenFE} OpenFE was introduced to fit the training set once and transform the validation and test sets to prevent information leakage.

\paragraph{CAAFE} We followed the original implementation of CAAFE. We changed the cross-validation setting in CAAFE to a single split of the training, validation, and test sets to ensure a fair comparison with MedFeat, which also enables finding the best threshold for classification. Other settings are kept as defaults in the original implementation. We removed the few-shot sample data in the original implementation.

\paragraph{FeatLLM} We followed the default implementation of FeatLLM and ran it with 5 ensemble queries. We performed missing value imputation for the entire dataset before execution, as required by the framework's configuration. Few-shot data samples were removed from the prompt. Since FeatLLM only splits the data into a train and test set and uses the default threshold of 0.5 for classification, reporting F1 scores on an imbalanced dataset will lead to biased comparison. Therefore, we did not report the F1 score for FeatLLM. Instead, we use AUPRC and AUROC and account for both ranking capability and performance on the rare class.

\paragraph{OCTree} We followed the original implementation of OCTree and ran for 3 steps. Since the released version is context-agnostic, we re-implemented the context-aware version of OCTree, following the original study configuration.

\paragraph{Ablation studies}
For the framework without model-awareness, we removed the structured prompt in the generation process. For the framework without island-based sampling, we chose the number of islands to be equal to one and the size of each island to be the number of features in the dataset. For the framework without feature importance, we sampled features uniformly to islands and removed the importance-related ranking in the generation process. For the ablation of memory and feedback, we retain no operation record or summary of MedFeat in each run.

\section{Dataset}
\label{app:data}

\subsection{Dataset and pre-processing}
Across these datasets, we consider multiple binary prediction tasks (e.g., mortality and readmission), yielding a total of seven tasks. For IORD data, we examine 24-hour mortality for the inpatient population and next-day discharge prediction. For MIMIC, we assess 24-hour mortality for ICU patients, the occurrence of heart failure within the next 30 days, and the readmission within the next 30 days. For HRS data, we examine the 10-year mortality risk and the occurrence of cognitive impairment in the next 10 years. The details of the dataset and tasks are provided in Table \ref{tab:dataset_stat}.

\begin{table*}[t]
\centering
\small

\renewcommand{\arraystretch}{1.15}

\resizebox{\textwidth}{!}{%
\begin{tabular}{@{}cccccc@{}}
\toprule
\textbf{Dataset} & \textbf{Task} & \textbf{N} & \textbf{$p$} & \textbf{Ratio (+:-)} & \textbf{Task Description} \\
\midrule
\multirow{2}{*}{IORD}
  & Mortality-Ward    & 92{,}237 & 31 & 1{:}313 & Predict whether the patient will die within 1 day. \\
  & Discharge         & 92{,}237 & 31 & 1{:}4   & Predict whether the patient will be discharged within 1 day. \\
\midrule
\multirow{3}{*}{MIMIC}
  & Mortality-ICU     & 90{,}257 & 16 & 1{:}47  & Predict whether the patient will die within 24 hours. \\
  & Readmission       & 74{,}044 & 17 & 1{:}4   & Predict whether the patient will be readmitted within 30 days. \\
  & Heart Failure     & 74{,}044 & 17 & 1{:}17  & Predict whether the patient will develop heart failure within 30 days. \\
\midrule
\multirow{2}{*}{HRS}
  & Mortality-10yrs   & 8{,}899  & 41 & 1{:}9   & Predict whether the participant will die within 10 years. \\
  & Cog-Impairment    & 8{,}899  & 41 & 1{:}2   & Predict whether the participant will be cognitively impaired within 10 years. \\
\bottomrule
\end{tabular}}
\caption{Statistics of the datasets and prediction tasks. $N$ denotes the number of samples, $p$ the number of features, and Ratio (+:-) the class balance between positive and negative samples.}
\label{tab:dataset_stat}
\end{table*}

\paragraph{Infections in the Oxfordshire Research Database} The Infections in Oxfordshire Research Database (IORD) is a large-scale electronic health record database covering patient demographics, mortality, hospital admissions, diagnostic codes, and vital signs. De-identified data were obtained from Oxford University Hospitals NHS Foundation Trust (OUH), consisting of 4 teaching hospitals in Oxfordshire, UK, collectively containing 1100 beds that serve approximately 1\% of the UK population and provide specialist regional referral services. We randomly selected time points across the study period and collected data from hospitalized patients in the previous 12 hours, following previous studies on similar tasks \cite{wei2024predicting}. Features used can be found in Table \ref{tab:iord}.

Data used in this section of the study are available from the Infections in Oxfordshire Research Database, subject to an application and research proposal meeting the ethical and governance requirements of the Database.

Deidentified data covering patient demographics, mortality, hospital admissions, diagnostic codes and vital signs were obtained from the Infections in Oxfordshire Research Database, which has approvals from the National Research Ethics Service South Central-Oxford C Research Ethics Committee (24/SC/0241), Health Research Authority and Confidentiality Advisory Group (19/CAG/0144) as a deidentified database without individual consent.

\begin{table}[h]
\centering

\resizebox{\columnwidth}{!}{%
\begin{tabular}{ll}
\toprule
\textbf{Feature} & \textbf{Type} \\
\midrule
Age & Numerical \\
Sex & Categorical \\
Index of multiple deprivation score & Numerical \\
Admission method code & Categorical \\
Hours in hospital after admission & Numerical \\
Number of unique medicines received last 24h & Numerical \\
\midrule
Heart Rate X hours ago $(X \in [0, 5])$ & Numerical \\
\midrule
Amount of oxygen delivered X hours ago $(X \in [0, 5])$ & Numerical \\
\midrule
Respiratory rate X hours ago $(X \in [0, 5])$ & Numerical \\
\midrule
oxygen saturation X hours ago $(X \in [0, 5])$ & Numerical \\
\bottomrule
\end{tabular}
}
\caption{Input features and data types for IORD data.}
\label{tab:iord}
\end{table}

\paragraph{Medical Information Mart for Intensive Care} We use data derived from the Medical Information Mart for Intensive Care IV (MIMIC-IV), a large, publicly available database containing de-identified electronic health records of patients admitted to intensive care units. MIMIC comprises detailed clinical information, including demographics, vital signs, laboratory measurements, medications, procedures, and outcomes, recorded at high temporal resolution throughout each hospital stay. Following the established pre-processing pipeline \cite{gupta2022extensive}, we collected the first 12 hours of patient data after their admission to the ICU when predicting mortality risk in the next 24 hours, and the last 12 hours of patient data before their discharge to predict the occurrence of heart failure in the next 30 days, and readmission in the next 30 days. Features used for each task are given in the following Table~\ref{tab:mimic_admission}, ~\ref{tab:mimic_discharge}.

MIMIC-IV is a publicly available database sourced from the electronic health records of the Beth Israel Deaconess Medical Center.

\begin{table}[h]
\centering

\resizebox{\columnwidth}{!}{%
\begin{tabular}{ll}
\toprule
\textbf{Feature} & \textbf{Type} \\
\midrule
Age at ICU admission & Numerical \\
Sex & Categorical \\
Admission type & Categorical \\
Admission location & Categorical \\
Insurance & Categorical \\
Language & Categorical \\
Marital status & Categorical \\
Race & Categorical \\
\midrule
Heart rate X hours after admission $(X \in \{0, 6, 12\})$ & Numerical \\
\midrule
Respiratory rate X hours after admission $(X \in \{0, 6, 12\})$ & Numerical \\
\midrule
Oxygen saturation X hours after admission $(X \in \{0, 6, 12\})$ & Numerical \\
\bottomrule
\end{tabular}
}
\caption{Input features and MIMIC data types for task Mortality-ICU}
\label{tab:mimic_admission}
\end{table}

\begin{table}[h]
\centering

\resizebox{\columnwidth}{!}{%
\begin{tabular}{ll}
\toprule
\textbf{Feature} & \textbf{Type} \\
\midrule
Age at ICU admission & Numerical \\
Sex & Categorical \\
Admission type & Categorical \\
Admission location & Categorical \\
Insurance & Categorical \\
Language & Categorical \\
Marital status & Categorical \\
Race & Categorical \\
Hours after admission & Numerical \\
\midrule
Heart rate X hours before discharge $(X \in \{0, 6, 12\})$ & Numerical \\
\midrule
Respiratory rate X hours before discharge $(X \in \{0, 6, 12\})$ & Numerical \\
\midrule
Oxygen saturation X hours before discharge $(X \in \{0, 6, 12\})$ & Numerical \\
\bottomrule
\end{tabular}
}
\caption{Input features and MIMIC data types for task Readmission and Heart Failure}
\label{tab:mimic_discharge}
\end{table}

\paragraph{Health and Retirement Study} The Health and Retirement Study (HRS) is a nationally representative longitudinal survey of adults aged 50 years and older in the United States. HRS collects rich information on respondents’ demographics, socioeconomic status, physical and mental health, health behaviors, and healthcare utilization, with follow-up interviews conducted every 2 years. Its longitudinal design enables the study of aging-related trajectories and long-term health outcomes, while its standardized survey instruments and broad population coverage make it a widely used resource in social science and health research. The HRS (Health and Retirement Study) is sponsored by the National Institute on Aging (grant number NIA U01AG009740) and is conducted by the University of Michigan. We followed previous studies to process the data to predict the 10-year mortality risk and cognitive impairment risk for study participants \cite{puterman2020predicting}. Features being used are given in the following Table~\ref{tab:hrs}. 

In this study, we use the RAND HRS Longitudinal File 2022, which contains public information from the HRS Core and Exit interview \cite{hrs_dataset}.

\begin{table}[]
\centering

\resizebox{\columnwidth}{!}{%
\begin{tabular}{ll}
\toprule
\textbf{Feature} & \textbf{Type} \\
\midrule
Respondent gender & Categorical \\
Respondent race & Categorical \\
Respondent years of education & Numerical \\
Respondent veteran status & Binary \\
\midrule
Respondent marital status at wave $w$ $(w \in \{8,9,10\})$ & Categorical \\
\midrule
Respondent age (years) at interview at wave $w$ $(w \in \{8,9,10\})$ & Numerical \\
\midrule
Self-report of health change at wave $w$ $(w \in \{8,9,10\})$ & Categorical \\
\midrule
CES-D depression score at wave $w$ $(w \in \{8,9,10\})$ & Numerical \\
\midrule
Self-reported body mass index (kg/m$^2$) at wave $w$ $(w \in \{8,9,10\})$ & Numerical \\
\midrule
Respondent currently smokes at wave $w$ $(w \in \{8,9,10\})$ & Binary \\
\midrule
Respondent ever drinks alcohol at wave $w$ $(w \in \{8,9,10\})$ & Binary \\
\midrule
High blood pressure since last interview at wave $w$ $(w \in \{8,9,10\})$ & Binary \\
\midrule
Diabetes since last interview at wave $w$ $(w \in \{8,9,10\})$ & Binary \\
\midrule
Hospital stay in previous two years at wave $w$ $(w \in \{8,9,10\})$ & Binary \\
\midrule
Total assets (cross-wave) at wave $w$ $(w \in \{8,9,10\})$ & Numerical \\
\midrule
Any difficulty in activities of daily living (ADL sum 0--5) at wave $w$ $(w \in \{8,9,10\})$ & Numerical \\
\bottomrule
\end{tabular}
}
\caption{Input features and HRS data types}
\label{tab:hrs}
\end{table}

\section{Additional Results}

\subsection{Significant test}
\label{app:significance}

We compare MedFeat against each baseline with a one-sided Wilcoxon
signed-rank test using the task as the unit of analysis
\cite{demvsar2006statistical}. We average over the two downstream learners within
each task and test each metric separately, giving 7 paired observations per
comparison. We apply Holm correction across the five comparisons. MedFeat is statistically significantly better than every baseline on AUROC.

\begin{table}[h]
\centering
\small
\begin{tabular}{lcc}
\toprule
\textbf{MedFeat vs.} & \textbf{mean $\Delta$} & \textbf{Holm-adj. $p$} \\
\midrule
Baseline & 1.71  & 0.039* \\
OpenFE   & 7.83  & 0.039* \\
CAAFE    & 1.61  & 0.039* \\
FeatLLM  & 20.18 & 0.039* \\
OCTree   & 1.34  & 0.039* \\
\bottomrule
\end{tabular}
\caption{One-sided Wilcoxon signed-rank test with Holm correction on AUROC
(MedFeat vs. baselines, 7 tasks, learners averaged within task).
mean $\Delta$ denotes the mean AUROC difference in favor of MedFeat.
*: $p < 0.05$.}
\label{tab:significance}
\end{table}

\subsection{Per-task AUPRC}
\label{app:auprc}

For completeness, Table~\ref{tab:auprc_lr} and Table~\ref{tab:auprc_xgb}
reports per-task AUPRC in the same setting as
Table~\ref{tab:main_results}.

\begin{table*}[t]
\centering
\small
\setlength{\tabcolsep}{4pt}
\begin{tabular}{lccccccc}
\toprule
Method & Cog-Impairment & Mortality-10yrs & Mortality-Ward & Discharge & Mortality-ICU & Heart Failure & Readmission \\
\midrule
Baseline & 56.3 {\scriptsize$\pm$1.9} & 25.3 {\scriptsize$\pm$0.8} & 3.6 {\scriptsize$\pm$0.4} & 31.3 {\scriptsize$\pm$0.8} & 11.6 {\scriptsize$\pm$1.0} & 7.4 {\scriptsize$\pm$0.2} & 18.8 {\scriptsize$\pm$0.4} \\
OpenFE   & 48.3 {\scriptsize$\pm$3.0} & 18.6 {\scriptsize$\pm$5.6} & 1.7 {\scriptsize$\pm$0.8} & 26.2 {\scriptsize$\pm$1.3} & 7.1 {\scriptsize$\pm$4.2} & \textbf{10.7} {\scriptsize$\pm$8.0} & 18.2 {\scriptsize$\pm$0.4} \\
CAAFE    & 56.3 {\scriptsize$\pm$1.9} & 25.4 {\scriptsize$\pm$0.9} & 3.6 {\scriptsize$\pm$0.4} & 31.3 {\scriptsize$\pm$0.8} & 11.8 {\scriptsize$\pm$1.1} & 7.6 {\scriptsize$\pm$0.4} & 19.0 {\scriptsize$\pm$0.4} \\
FeatLLM  & 34.2 {\scriptsize$\pm$2.6} & 12.8 {\scriptsize$\pm$3.0} & 2.1 {\scriptsize$\pm$1.3} & 17.4 {\scriptsize$\pm$0.2} & 2.1 {\scriptsize$\pm$0.0} & 4.5 {\scriptsize$\pm$0.0} & 15.7 {\scriptsize$\pm$0.0} \\
OCTree   & 56.3 {\scriptsize$\pm$1.9} & 25.0 {\scriptsize$\pm$0.9} & 3.4 {\scriptsize$\pm$0.4} & 31.3 {\scriptsize$\pm$0.8} & 12.8 {\scriptsize$\pm$1.1} & 7.6 {\scriptsize$\pm$0.3} & 19.0 {\scriptsize$\pm$0.3} \\
\textsc{MedFeat} & \textbf{58.8} {\scriptsize$\pm$1.6} & \textbf{26.3} {\scriptsize$\pm$1.1} & \textbf{3.9} {\scriptsize$\pm$0.7} & \textbf{34.6} {\scriptsize$\pm$0.7} & \textbf{13.6} {\scriptsize$\pm$1.1} & 7.8 {\scriptsize$\pm$0.4} & \textbf{19.9} {\scriptsize$\pm$0.5} \\
\bottomrule
\end{tabular}
\caption{Per-task AUPRC (\%) under logistic regression averaged over
5 random splits. Bold indicates the best method per task.}
\label{tab:auprc_lr}
\end{table*}

\begin{table*}[t]
\centering
\small
\setlength{\tabcolsep}{4pt}
\begin{tabular}{lccccccc}
\toprule
Method & Cog-Impairment & Mortality-10yrs & Mortality-Ward & Discharge & Mortality-ICU & Heart Failure & Readmission \\
\midrule
Baseline & 54.9 {\scriptsize$\pm$1.9} & 19.6 {\scriptsize$\pm$2.5} & 1.7 {\scriptsize$\pm$0.3} & 37.0 {\scriptsize$\pm$0.6} & 18.1 {\scriptsize$\pm$2.7} & 8.2 {\scriptsize$\pm$0.6} & 20.3 {\scriptsize$\pm$0.5} \\
OpenFE   & 55.2 {\scriptsize$\pm$1.3} & 20.8 {\scriptsize$\pm$2.6} & \textbf{2.6} {\scriptsize$\pm$1.9} & 34.8 {\scriptsize$\pm$0.9} & 12.1 {\scriptsize$\pm$0.9} & 7.4 {\scriptsize$\pm$0.4} & 19.7 {\scriptsize$\pm$0.3} \\
CAAFE    & 55.1 {\scriptsize$\pm$1.6} & 19.7 {\scriptsize$\pm$2.2} & 2.2 {\scriptsize$\pm$1.2} & 37.1 {\scriptsize$\pm$0.7} & 16.2 {\scriptsize$\pm$2.3} & 8.3 {\scriptsize$\pm$0.5} & 20.6 {\scriptsize$\pm$0.6} \\
FeatLLM  & 31.0 {\scriptsize$\pm$1.1} & 11.5 {\scriptsize$\pm$2.0} & 0.3 {\scriptsize$\pm$0.0} & 17.3 {\scriptsize$\pm$0.0} & 2.1 {\scriptsize$\pm$0.0} & 4.5 {\scriptsize$\pm$0.0} & 15.7 {\scriptsize$\pm$0.0} \\
OCTree   & 55.5 {\scriptsize$\pm$1.3} & 19.1 {\scriptsize$\pm$2.4} & 1.9 {\scriptsize$\pm$0.5} & 37.0 {\scriptsize$\pm$0.5} & 18.0 {\scriptsize$\pm$2.1} & 8.3 {\scriptsize$\pm$0.2} & 20.4 {\scriptsize$\pm$0.3} \\
\textsc{MedFeat} & \textbf{57.2} {\scriptsize$\pm$0.9} & \textbf{21.7} {\scriptsize$\pm$1.5} & 2.4 {\scriptsize$\pm$0.6} & \textbf{37.4} {\scriptsize$\pm$0.5} & \textbf{20.3} {\scriptsize$\pm$1.9} & \textbf{8.9} {\scriptsize$\pm$0.4} & \textbf{20.9} {\scriptsize$\pm$0.6} \\
\bottomrule
\end{tabular}
\caption{Per-task AUPRC (\%) under XGBoost averaged over 5 random
splits. Bold indicates the best method per task.}
\label{tab:auprc_xgb}
\end{table*}

\subsection{Token usage and computational efficiency}

MedFeat consumes approximately 9,900 tokens per task on average, higher than OCTree and CAAFE, reflecting the richer prompt context arising from importance signals, model-aware instructions, and feedback memory. Nonetheless, this budget remains well within practical limits and is lower than that of FeatLLM ($\sim$11,500), which achieves the weakest performance despite the highest cost. MedFeat also exhibits the lowest standard deviation in token usage across tasks ($\sim$620, compared with $\sim$860 for FeatLLM and $\sim$1,350 for OCTree), ensuring predictable expenditure across diverse clinical settings.

Several methodological considerations are important when interpreting these token counts, as the comparison is not entirely aligned and possibly understates MedFeat's relative efficiency. First, the OCTree numbers reported here correspond to our context-aware re-implementation; the original released version of OCTree is context-agnostic and would consume substantially fewer tokens, as it omits the dataset- and task-specific context that grounds clinically meaningful generation. Because the original context-aware variant was not released, our re-implementation may not fully recover all components of the original design and could therefore underestimate OCTree's true token footprint when deployed in a context-aware setting.

Second, the CAAFE and FeatLLM token counts reported here are lower than in their original implementations because we removed the few-shot raw patient samples from their prompts, which increases token usage considerably. We stripped these few-shot samples for a principled reason: in clinical deployment, transmitting patient-level information to online LLMs is prohibited under privacy and data governance constraints. Without this modification, CAAFE and FeatLLM would consume more tokens than the values reported, and the gap between them and MedFeat would narrow or invert. In other words, the relatively modest token counts of CAAFE and FeatLLM in our comparison are a consequence of a privacy-preserving adaptation we applied uniformly, not a property of their native designs.

\begin{figure}[]
  \centering
  \hspace{-0.05\columnwidth}
  \includegraphics[width=1\columnwidth]{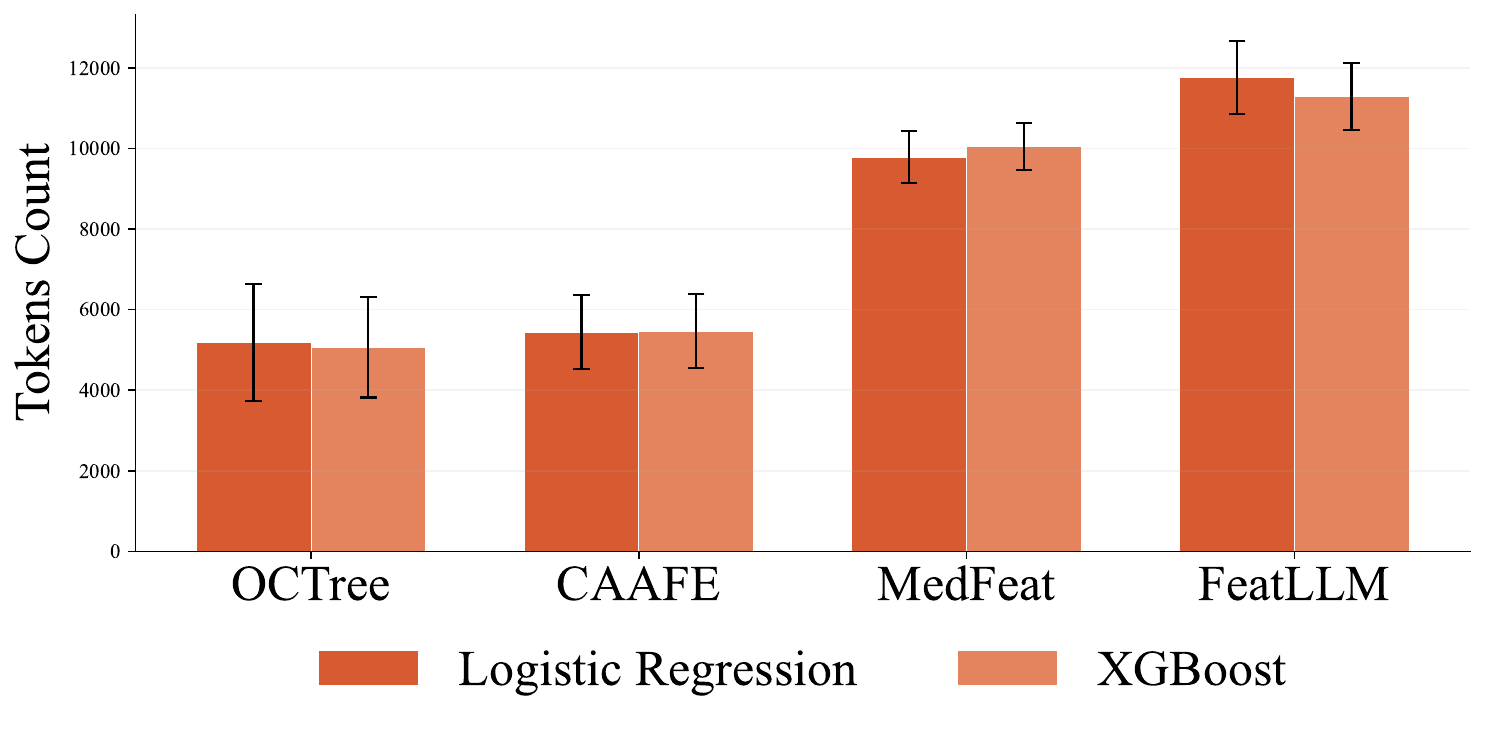}
  \caption{Total token usage of different methods. Mean and std of the usage are averaged across 7 tasks.}
  \label{fig:token}
\end{figure}

It is also worth situating these numbers in absolute economic terms. At current commercial LLM pricing, roughly 9,900 tokens per task corresponds to a cost on the order of cents, even with premium models. Because feature engineering is performed once during model development rather than at every inference call, this expenditure is amortized across the entire downstream deployment lifetime of the resulting model. Relative to the engineering hours typically required for manual clinical feature construction, or to the downstream value of even a modest improvement in mortality or readmission prediction, the marginal monetary cost of MedFeat's larger prompts is insignificant and should not be a practical barrier to adoption.

Taken together, MedFeat's increased token consumption represents a deliberate and well-justified trade-off. The additional tokens encode signals that no baseline provides in privacy-preserving form, and they translate directly into statistically significant gains (Table \ref{tab:main_results}, Table \ref{tab:significance}). On a cost-per-unit-improvement basis, MedFeat remains highly favorable: it sits below the most expensive baseline (FeatLLM) while substantially outperforming it, and the modest premium over OCTree and CAAFE buys consistent improvements that those leaner prompts demonstrably cannot achieve. Combined with MedFeat's tighter token-count variance across tasks, this profile makes the framework predictable to budget for in real-world clinical pipelines, where reliable resource estimation is often as important as raw efficiency.

\subsection{Clinical utility}

We examine the
patient-level confusion matrix on Mortality-Ward, our most imbalanced task
(1:313), under XGBoost to investigate the clinical meaningfulness of MedFeat under smaller absolute F1 values.

\begin{table}[h]
\centering
\small
\begin{tabular}{lccc}
\toprule
\textbf{Metric} & \textbf{Baseline} & \textbf{MedFeat} & \textbf{$\Delta$} \\
\midrule
Deaths caught (TP)  & 36.0   & 39.2   & +3.2 \\
Deaths missed (FN)  & 23.0   & 19.8   & --3.2 \\
False alarms (FP)   & 5225.6 & 4318.2 & --907.4 \\
F1                  & 1.4    & 1.8    & +0.4 \\
\bottomrule
\end{tabular}
\caption{Confusion matrix analysis on Mortality-Ward (XGBoost,
$N = 18{,}447$ test episodes), averaged over 5 random splits at the
validation-selected operating point.}
\label{tab:confusion_ward}
\end{table}

As shown in Table~\ref{tab:confusion_ward}, the F1 change from 1.4 to 1.8
corresponds to identifying 3.2 additional deaths (an 8.9\% relative increase
in sensitivity) while simultaneously removing 907.4 false alarms (a 17.4\%
relative reduction). In deployed clinical early-warning systems, raising sensitivity
while reducing alert fatigue is precisely the direction of interest.

Demonstrating overall clinical utility would benefit from matched-operating-point
comparisons, calibration assessment, and decision-curve analysis, which we
leave to future work.

\begin{table*}[]
\centering
\small
\setlength{\tabcolsep}{3pt}
\begin{tabular}{lcccccc}
\toprule
\multirow{2}{*}{\textbf{Method}} & \multicolumn{2}{c}{\textbf{Mortality-Ward}} & \multicolumn{2}{c}{\textbf{Mortality-10yrs}} & \multicolumn{2}{c}{\textbf{Mortality-ICU}} \\
\cmidrule(lr){2-3}\cmidrule(lr){4-5}\cmidrule(lr){6-7}
 & F1 & AUROC & F1 & AUROC & F1 & AUROC \\
\midrule
Raw RF            & 2.6\scriptsize{$\pm$0.3} & 72.2\scriptsize{$\pm$3.8} & 28.1\scriptsize{$\pm$2.1} & 74.0\scriptsize{$\pm$2.2} & 10.9\scriptsize{$\pm$0.4} & 79.4\scriptsize{$\pm$1.5} \\
MedFeat + GI      & 2.7\scriptsize{$\pm$0.3} & 73.8\scriptsize{$\pm$3.2} & 29.4\scriptsize{$\pm$3.1} & 74.6\scriptsize{$\pm$2.4} & 11.4\scriptsize{$\pm$1.9} & 79.7\scriptsize{$\pm$1.1} \\
MedFeat + SHAP    & \textbf{2.8}\scriptsize{$\pm$0.3} & \textbf{74.5}\scriptsize{$\pm$3.3} & \textbf{29.8}\scriptsize{$\pm$1.4} & \textbf{74.7}\scriptsize{$\pm$1.9} & \textbf{11.5}\scriptsize{$\pm$0.3} & \textbf{79.9}\scriptsize{$\pm$0.5} \\
\bottomrule
\end{tabular}
\caption{Performance (\%) of MedFeat with a Random Forest downstream model
under Gini importance (GI) and SHAP attribution. Mean and standard deviation
over 5 random splits.}
\label{tab:rf_gini}
\end{table*}

\subsection{MedFeat generalizes across learners and attribution signals}
\label{app:generality}

The main experiments pair SHAP attribution with logistic regression and
XGBoost. To verify that MedFeat is not tied to either choice, we vary the two
axes separately: (Q1) does MedFeat generalize to other downstream learners,
including a neural one, and (Q2) is SHAP a better attribution signal than a
learner-native signal?

\paragraph{Neural learner.}
We run MedFeat with a multilayer perceptron (MLP) using two hidden layers of
sizes 128 and 64 with ReLU activations, the Adam optimizer with $\ell_2$
regularization, a learning rate of $10^{-3}$, a batch size of 256, and up to
300 epochs with early stopping. All other settings follow the common setup.
Table~\ref{tab:mlp} shows that MedFeat improves the MLP baseline on all seven
tasks for both F1 and AUROC, raising average F1 from 24.9 to 26.4 and average
AUROC from 69.0 to 72.2. The gains are largest on Mortality-10yrs
(AUROC $60.7 \rightarrow 73.3$) and Mortality-Ward
(AUROC $72.8 \rightarrow 79.2$), and modest on tasks such as Readmission and
Heart Failure.

\begin{table}[]
\centering
\small
\setlength{\tabcolsep}{3pt}
\begin{tabular}{lcccc}
\toprule
\multirow{2}{*}{Task} & \multicolumn{2}{c}{Baseline} & \multicolumn{2}{c}{\textsc{MedFeat}} \\
\cmidrule(lr){2-3}\cmidrule(lr){4-5}
 & F1 & AUROC & F1 & AUROC \\
\midrule
Mortality-ICU   & 9.6\tiny$\pm$1.0  & 78.4\tiny$\pm$2.7 & \textbf{10.4}\tiny$\pm$1.0 & \textbf{80.3}\tiny$\pm$1.0 \\
Mortality-10yrs & 19.5\tiny$\pm$1.3 & 60.7\tiny$\pm$1.8 & \textbf{27.7}\tiny$\pm$2.5 & \textbf{73.3}\tiny$\pm$3.7 \\
Heart Failure   & 13.8\tiny$\pm$0.5 & 65.5\tiny$\pm$1.4 & \textbf{13.9}\tiny$\pm$0.6 & \textbf{66.2}\tiny$\pm$1.4 \\
Mortality-Ward  & 1.8\tiny$\pm$0.5  & 72.8\tiny$\pm$1.7 & \textbf{2.0}\tiny$\pm$0.5  & \textbf{79.2}\tiny$\pm$3.8 \\
Cog-Impairment  & 58.9\tiny$\pm$1.3 & 76.4\tiny$\pm$1.4 & \textbf{59.4}\tiny$\pm$1.3 & \textbf{77.1}\tiny$\pm$1.5 \\
Discharge       & 39.9\tiny$\pm$0.6 & 70.9\tiny$\pm$0.6 & \textbf{40.3}\tiny$\pm$0.5 & \textbf{71.4}\tiny$\pm$0.5 \\
Readmission     & 30.6\tiny$\pm$1.0 & 58.0\tiny$\pm$0.6 & \textbf{31.0}\tiny$\pm$0.7 & \textbf{58.2}\tiny$\pm$0.5 \\
\midrule
Average         & 24.9 & 69.0 & \textbf{26.4} & \textbf{72.2} \\
\bottomrule
\end{tabular}
\caption{Performance (\%) of MedFeat with an MLP downstream learner across the
7 clinical prediction tasks (mean and std over 5 random splits).
MedFeat improves over the raw-feature baseline on every task and both metrics.}
\label{tab:mlp}
\end{table}

\paragraph{Alternative attribution signal.}
We further run MedFeat with a Random Forest (RF) downstream model under two
attribution signals, Gini Index (GI) and SHAP, on three mortality tasks
(Table~\ref{tab:rf_gini}). MedFeat improves over the raw RF baseline on all
three tasks under \emph{both} signals. MedFeat + SHAP nonetheless
matches or exceeds MedFeat + GI on every task, with consistently smaller
variance (e.g.\ $\pm$0.5 vs.\ $\pm$1.1 AUROC on Mortality-ICU). This supports
our default choice: GI is computed on the training data and is biased across
mixed feature types and cardinalities \cite{strobl2007bias}, which
destabilizes the ranking across splits, whereas SHAP derives importance from
model predictions and yields a more stable signal
\cite{lundberg2020local}.

Taken together, MedFeat improves four learner families with distinct inductive
biases (logistic regression, XGBoost, MLP, Random Forest) and remains
effective under two different attribution signals.

\begin{table}[]
\centering
\small
\begin{tabular}{lccccc}
\toprule
\textbf{Task} & \textbf{Acc.} & \textbf{Rej.} & \textbf{Inv.} & \textbf{\% Runs $\uparrow$} \\
\midrule
Cog-Impairment   & 6.0  & 4.8 & 0.1 & 100\% \\
Mortality-10yrs  & 4.1  & 5.3 & 0.2 & 90\%  \\
Mortality-Ward   & 3.4  & 4.2 & 0.0 & 80\%  \\
Discharge        & 4.5  & 4.0 & 0.3 & 90\%  \\
Mortality-ICU    & 4.3  & 3.7 & 0.2 & 100\% \\
Heart Failure    & 6.2  & 4.1 & 0.1 & 100\% \\
Readmission      & 15.1 & 5.7 & 0.2 & 100\% \\
\bottomrule
\end{tabular}
\caption{Accepted, rejected, and invalid feature transformations per task,
averaged over 5 splits and 2 downstream models. \% Runs $\uparrow$ denotes the
proportion of runs in which test AUROC improves over the raw baseline.}
\label{tab:feature_counts}
\end{table}

\begin{table*}[h]
\centering
\small
\setlength{\tabcolsep}{4pt}
\begin{tabular}{@{}llll@{}}
\toprule
\textbf{Task} & \textbf{Pattern} & \textbf{Example} & \textbf{Ref.} \\
\midrule
Mortality-Ward   & Clinical scores        & Early Warning Score (EWS)                        & \cite{williams2022national} \\
Mortality-10yrs  & Behavioral risk       & Alcohol $\times$ depression                      & \cite{yan2025heavy} \\
Heart Failure    & Demographics + vitals  & Age $\times$ heart rate change                   & \cite{ostrowska2024association} \\
Readmission      & Vital sign instability & Var(SpO$_2$) $\times$ Var(respiratory rate)      & \cite{nguyen2017vital} \\
\bottomrule
\end{tabular}
\caption{Representative accepted feature patterns, with example
transformations and supporting clinical literature.}
\label{tab:feature_patterns}
\end{table*}

\begin{table}[h]
\centering
\small
\begin{tabular}{lccc}
\toprule
\textbf{Method} & \textbf{F1} & \textbf{AUROC} & \textbf{AUPRC} \\
\midrule
CAAFE  & 25.1 $\pm$ 0.7 & 70.6 $\pm$ 1.2 & 22.4 $\pm$ 1.0 \\
FeatLLM  & --             & 58.9 $\pm$ 4.4 & 15.1 $\pm$ 1.8 \\
\textsc{MedFeat}   & \textbf{26.2 $\pm$ 1.0} & \textbf{71.5 $\pm$ 1.1} & \textbf{23.9 $\pm$ 1.0} \\
\bottomrule
\end{tabular}
\caption{MedFeat versus the original few-shot implementations of CAAFE and
FeatLLM under the Qwen-3.6-27B backbone. Mean $\pm$ std over 7 tasks and 2
downstream models.}
\label{tab:fewshot}
\end{table}

\subsection{Feature patterns summary}
\label{app:feature_report}

Table~\ref{tab:feature_counts} reports the number of accepted, rejected, and
invalid feature transformations per task, averaged over 5 random splits and both
downstream learners. A transformation is counted as \emph{invalid} when the
generated code fails execution or returns non-numeric outputs after three
retries, and as \emph{accepted} only when it
improves validation performance under the tolerance threshold. The held-out test set is evaluated once at the end
of the run with all accepted features. Invalid generations are rare
($\leq 0.3$ per task). Acceptance rates vary with task
difficulty, where Readmission accumulates the largest number of accepted features
(15.1), while Mortality-Ward, the most severely imbalanced task, accepts the
fewest (3.4). MedFeat improves test AUROC in 80-100\% of runs across all tasks.

Additionally, we summarize some representative transformation patterns in Table~\ref{tab:feature_patterns}. All listed patterns correspond to
established clinical evidence, and we cite supporting literature where one
exists.

\subsection{MedFeat's advantage persists compared with few-shot baselines}
\label{app:fewshot}

Our main experiments remove few-shot raw patient rows from the prompts of
CAAFE and FeatLLM, since transmitting patient-level records to external APIs
is prohibited in clinical deployment. To verify that this privacy-preserving adaptation does
not drive our conclusions, we additionally run the \emph{fully original}
few-shot implementations of CAAFE and FeatLLM on all tasks, using the
open-weight Qwen-3.6-27B backbone, which is hosted locally.

Table~\ref{tab:fewshot} shows that MedFeat retains its advantage on every
metric even under this most generous setting for the baselines. For CAAFE,
restoring few-shot samples yields only a marginal change, indicating that its benefit comes
mainly from dataset context rather than from raw samples. For FeatLLM,
few-shot samples improve performance meaningfully (AUPRC $12.3 \rightarrow
15.1$), yet it remains the weakest method overall.

\subsection{Comparison with a handcrafted temporal-aggregation baseline}
\label{app:handcrafted}

We implement a non-LLM handcrafted baseline for comprehensive evaluation. For every repeated measurement in a
dataset, we generate mean, min, max, std, range, slope, and delta features,
and then apply SHAP-based selection to retain the top-ranked subset, matching
MedFeat's feature budget (Table~\ref{tab:feature_counts}). 

\begin{table*}[t]
\centering
\small
\setlength{\tabcolsep}{4pt}
\begin{tabular}{lccccccc}
\toprule
 & Mortality-ICU & Mortality-10yrs & Heart Failure & Mortality-Ward & Cog-Impairment & Discharge & Readmission \\
\midrule
\# features & 21 & 84 & 21 & 28 & 84 & 28 & 21 \\
\midrule
XGB F1  & 10.1$\pm$0.5 & 25.2$\pm$2.1 & 13.7$\pm$0.4$^{\ddagger}$ & 1.5$\pm$0.3 & 56.8$\pm$1.3 & 40.5$\pm$0.5 & 29.5$\pm$1.0$^{\dagger}$ \\
XGB AUC & 76.5$\pm$1.4 & 68.8$\pm$2.1 & 65.5$\pm$1.9$^{\dagger}$ & 74.6$\pm$2.7 & 74.3$\pm$1.4 & 71.2$\pm$0.6 & 56.3$\pm$1.0$^{\dagger}$ \\
LR F1   & 9.2$\pm$0.8 & 27.5$\pm$3.7$^{\dagger}$ & 14.4$\pm$0.2$^{\ddagger}$ & 2.3$\pm$0.7 & 57.5$\pm$1.1 & 38.5$\pm$0.5 & 30.1$\pm$0.8$^{\ddagger}$ \\
LR AUC  & 77.7$\pm$1.2 & 74.9$\pm$1.8$^{\dagger}$ & 68.1$\pm$1.1$^{\ddagger}$ & 84.0$\pm$1.4 & 75.7$\pm$1.0 & 69.6$\pm$0.2 & 57.4$\pm$0.8$^{\dagger}$ \\
\bottomrule
\end{tabular}
\caption{Handcrafted temporal-aggregation baseline with SHAP-based selection
(mean and std over 5 random splits, default hyperparameters).
$\dagger$: worse than the raw baseline in Table~\ref{tab:main_results};
$\ddagger$: outperforms MedFeat.}
\label{tab:handcrafted}
\end{table*}

MedFeat outperforms this baseline on 24/28 (86\%)
combinations (Table~\ref{tab:handcrafted}). On the 2 tasks where the handcrafted baseline beats MedFeat on F1, it falls below the raw baseline on AUROC. Overall, the handcrafted baseline falls
below the raw feature set in 6/28 (21\%) of cases.

This confirms two key advantages of MedFeat. First, targeted generation: like OpenFE, the
handcrafted pipeline is an undirected generate-and-filter procedure that
expands 21-84 candidate features before selection, which introduces noise on
imbalanced tasks; MedFeat proposes a small number of candidates directly and
validates each one. Second, cross-variable reasoning: SHAP filtering over
predefined operators can only retain simple per-variable aggregates
(e.g.\ maximum heart rate), whereas MedFeat create transformations that
couple static covariates with dynamic trends, which predefined operator libraries cannot express.

\begin{table}[h]
\centering
\small
\setlength{\tabcolsep}{4pt}
\begin{tabular}{lcccc}
\toprule
$\beta$ & Mortality-Ward & Discharge & Heart Failure & Mean \\
\midrule
0.000 & +0.52 & +0.66 & +0.28 & +0.49 \\
0.005 & +0.38 & +0.66 & +0.38 & +0.47 \\
0.010 & +0.40 & +0.90 & +0.70 & +0.67 \\
0.020 & +0.28 & +0.74 & +0.76 & +0.59 \\
\bottomrule
\end{tabular}
\caption{Sensitivity of MedFeat to the tolerance threshold $\beta$. Values are absolute
F1 improvements (percentage points) over the raw baseline under XGBoost.}
\label{tab:beta}
\end{table}

\subsection{Sensitivity to the tolerance threshold $\beta$}
\label{app:beta}

We vary $\beta \in \{0, 0.005, 0.01, 0.02\}$ on three
representative tasks spanning different imbalance regimes (Mortality-Ward, Discharge,
Heart Failure) with XGBoost, and report the absolute F1 gain over the raw baseline in
Table~\ref{tab:beta}.

The mean gain varies by only
0.20pp across the sweep, with $\beta = 0.01$ performing best on average. The trend is mildly task-dependent: a strict threshold ($\beta = 0$)
favors the most imbalanced task (Mortality-Ward), where admitting marginal features is most likely to propagate noise, whereas a looser
threshold benefits tasks like Heart Failure.

\section{Scalability}
\label{sec:scalability}

Here we discuss
how MedFeat behaves as the feature dimensionality $M$ grows, as in
high-dimensional EHR settings with hundreds or thousands of diagnosis,
medication, laboratory, and procedure variables. MedFeat is designed
such that its core components do not grow with $M$.

\paragraph{Stable prompt size.} Each LLM call receives only an island
of $m$ features, so prompt length is
governed by $m$ and the memory bank rather than by $M$. Prompt size is
therefore approximately fixed as the table widens.

\paragraph{Importance-weighted sampling becomes more valuable as $M$
grows.} With many mostly low-signal variables, sampling according to
$p_t(i)$ concentrates the search on predictive regions of the feature
space. By contrast, exhaustive operator-based methods
\citep{horn2019autofeat, zhang2023openfe} expand combinatorially with
$M$, which is the failure mode motivated
and observed in the instability of OpenFE in Table~\ref{tab:main_results}.

Therefore, MedFeat has architectural properties that scale with data dimensions. We note that demonstrating MedFeat on high-dimensional data would further
strengthen the framework.

\section{Privacy}

Clinical deployment of LLM-assisted pipelines is constrained by institutional regulations that prohibit transmitting identifiable patient records to third-party APIs. Even de-identified rows remain vulnerable to re-identification, particularly for rare phenotypes. Prior LLM-based methods are poorly suited to this setting: CAAFE and FeatLLM include few-shot patient rows in the prompt, and the context-aware variant of OCTree exposes value distributions through decision-tree summaries.

MedFeat transmits only two types of information, which differ fundamentally in their privacy implications. The first is schema metadata: feature names and types. This contains no patient information and corresponds to what is routinely published in data dictionaries and methods sections. The second is aggregated SHAP importance scores computed locally on the validation set; these are population-level summary statistics over all rows, with no row-level reconstruction possible from the scalar values transmitted, and are equivalent to the standard covariate importance reported in clinical studies. No raw records, value distributions, or outcome-conditioned summaries are sent to LLMs. We note that institutions with stricter requirements can host open-source LLM backbones locally; MedFeat's effectiveness across backbones (Table \ref{tab:cross_llm_comparison}), including the smaller Qwen-27B, supports this deployment path.

\section{Software, Computation Resources and Reproducibility}
All experiments were conducted using Python 3.11. LLM queries were performed via the OpenAI API and Google Cloud. Please refer to the corresponding sections for data availability (Appendix \ref{app:data}). Code required to reproduce our experiments can be found in: \url{https://github.com/zizhengzhang1/MedFeat}.

\end{document}